\documentclass{article}

\PassOptionsToPackage{numbers,sort&compress}{natbib}
\usepackage{report}
\setcitestyle{numbers,square,comma}

\usepackage[utf8]{inputenc}
\usepackage[T1]{fontenc}
\usepackage{hyperref}
\usepackage{url}
\usepackage{booktabs}
\usepackage{amsfonts}
\usepackage{amsmath}
\usepackage{amssymb}
\usepackage{microtype}
\usepackage{graphicx}
\usepackage{xcolor}
\usepackage{multirow}
\usepackage{subcaption}
\usepackage[most]{tcolorbox}
\usepackage{colortbl}
\usepackage{pifont}
\usepackage{capt-of}
\usepackage{needspace}
\usepackage{nicefrac}
\usepackage{wrapfig}
\usepackage{enumitem}
\usepackage{textcomp}
\usepackage{placeins}
\usepackage{fontawesome5}

\definecolor{xhsPink}{RGB}{232,190,190}
\definecolor{xhsPinkLight}{RGB}{253,245,245}
\newtcolorbox{abstractbox}{
    colback=xhsPinkLight, colframe=xhsPink,
    boxrule=1pt, arc=4mm,
    left=10pt, right=10pt, top=8pt, bottom=8pt,
    opacityback=0.95
}

\definecolor{todored}{HTML}{C62828}

\definecolor{macaronpink}{HTML}{FCE4EC}
\definecolor{macaronmint}{HTML}{E0F2E9}
\definecolor{macaronlemon}{HTML}{FFF8E1}
\definecolor{rowtintA}{HTML}{FAFAFA}
\definecolor{rowtintB}{HTML}{F1F4F8}

\definecolor{promptboxbg}{HTML}{FDF1F5}
\definecolor{promptboxframe}{HTML}{E8B4C5}
\definecolor{promptboxtitlebg}{HTML}{F4C8D5}
\definecolor{promptboxtitlefg}{HTML}{6E2940}
\definecolor{configboxbg}{HTML}{F2F9EE}
\definecolor{configboxframe}{HTML}{B5D6A7}
\definecolor{configboxtitlebg}{HTML}{CDE6BF}
\definecolor{configboxtitlefg}{HTML}{2E5A2E}

\newtcolorbox{promptbox}[1]{%
  enhanced,
  breakable,
  colback=promptboxbg, colframe=promptboxframe,
  boxrule=0.4pt, arc=2.5pt, outer arc=2.5pt,
  left=8pt, right=8pt, top=6pt, bottom=6pt,
  fonttitle=\scriptsize\bfseries\sffamily,
  coltitle=promptboxtitlefg,
  colbacktitle=promptboxtitlebg,
  attach boxed title to top left={xshift=10pt, yshift=-7pt},
  boxed title style={
    colback=promptboxtitlebg, colframe=promptboxtitlebg,
    boxrule=0pt, arc=2pt, outer arc=2pt,
    left=6pt, right=6pt, top=2pt, bottom=2pt,
  },
  title={#1},
  top=10pt,
  fontupper=\scriptsize\ttfamily,
  before skip=8pt, after skip=8pt,
}
\newtcolorbox{promptboxbreak}[1]{%
  enhanced,
  breakable,
  colback=promptboxbg, colframe=promptboxframe,
  boxrule=0.4pt, arc=2.5pt, outer arc=2.5pt,
  left=8pt, right=8pt, top=6pt, bottom=6pt,
  fonttitle=\scriptsize\bfseries\sffamily,
  coltitle=promptboxtitlefg,
  colbacktitle=promptboxtitlebg,
  attach boxed title to top left={xshift=10pt, yshift=-7pt},
  boxed title style={
    colback=promptboxtitlebg, colframe=promptboxtitlebg,
    boxrule=0pt, arc=2pt, outer arc=2pt,
    left=6pt, right=6pt, top=2pt, bottom=2pt,
  },
  title={#1},
  top=10pt,
  fontupper=\scriptsize\ttfamily,
  before skip=8pt, after skip=8pt,
}
\newtcolorbox{configbox}[1]{%
  enhanced,
  breakable,
  colback=configboxbg, colframe=configboxframe,
  boxrule=0.4pt, arc=2.5pt, outer arc=2.5pt,
  left=8pt, right=8pt, top=6pt, bottom=6pt,
  fonttitle=\scriptsize\bfseries\sffamily,
  coltitle=configboxtitlefg,
  colbacktitle=configboxtitlebg,
  attach boxed title to top left={xshift=10pt, yshift=-7pt},
  boxed title style={
    colback=configboxtitlebg, colframe=configboxtitlebg,
    boxrule=0pt, arc=2pt, outer arc=2pt,
    left=6pt, right=6pt, top=2pt, bottom=2pt,
  },
  title={#1},
  top=10pt,
  fontupper=\scriptsize,
  before skip=8pt, after skip=8pt,
}

\title{VibeSearchBench: Benchmarking Long-horizon Proactive Search in the Wild}

\author{
\textbf{Xiaohongshu Dots Studio \& Unipat AI}
}

\date{}

\hypersetup{
	pdftitle={VibeSearchBench: Benchmarking Long-horizon Proactive Search in the Wild},
	pdfkeywords={search benchmark, LLM agents, information seeking},
}

\begin{document}
\thispagestyle{firstpage}

\begin{abstractbox}
\begin{center}
{\LARGE\bfseries \thetitle\par}
\vspace{8pt}
{\normalsize \theauthor\par}
\vspace{4pt}
{\small\href{https://vibebench.github.io/VibeSearchBench.github.io/}{\textcolor[RGB]{160,60,60}{\texttt{https://vibebench.github.io/VibeSearchBench}}}}
\end{center}
\vspace{6pt}
\noindent{\large\bfseries Abstract}\par\vspace{4pt}
\noindent
LLM-based agents score well on search benchmarks, yet real users consistently find results unsatisfying, revealing a persistent \emph{evaluation--experience gap}.
We attribute this gap to existing benchmarks' reliance on over-specified queries, single-turn interactions, and fixed-schema evaluation, none of which reflect real search behavior where users and agents collaboratively refine vague intent through multi-turn dialogue.
We term this paradigm \textbf{VibeSearch} and introduce \textbf{VibeSearchBench}, a benchmark comprising 200 manually curated bilingual (Chinese and English) tasks across 20 domains, split into VibeSearch-Pro (professional) and VibeSearch-Daily (daily-life) subsets.
Each task pairs a user persona with a schema-free ground-truth knowledge graph, and is evaluated through a progressive-disclosure user simulator and a graph-matching evaluation framework.
We benchmark seven frontier models under both the ReAct framework and the OpenClaw agent harness.
Results show that all models remain substantially inadequate for VibeSearch (best F1: 30.30), highlighting the need for fundamental advances in long-context reasoning, proactive intent elicitation, and structured knowledge construction.
\end{abstractbox}
\section{Introduction}

Large Language Model-based AI agents have emerged as powerful search specialists~\cite{openai-dr,tongyideepresearchteam2025tongyideepresearchtechnicalreport}, capable of navigating complex real-world web environments through hundreds of tool-calling to find the proverbial ``needle in a haystack.''
Yet a persistent evaluation--experience gap remains: frontier models achieve ever-higher scores on benchmarks such as BrowseComp~\cite{wei2025browsecompsimplechallengingbenchmark} and WideSearch~\cite{wong2025widesearchbenchmarkingagenticbroad}, while real end-users continue to report that the results are ``off-topic,'' or ``don't understand me.''

A fundamental reason is the mismatch between how benchmarks frame search tasks and how users actually search.
In practice, most users do not, and indeed cannot, fully articulate their information needs upfront.
A realistic search session unfolds as an iterative user-agent interaction: (\textsc{User}) a vague query \textrightarrow (\textsc{Agent}) partial results and clarification \textrightarrow (\textsc{User}) expresses emerging preferences and needs  \textrightarrow (\textsc{Agent}) adjusts its search direction \textrightarrow (user-agent interaction) ... \textrightarrow the information need gradually converges into a concrete solution.
We term this class of tasks \textbf{VibeSearch}.

\begin{figure}[t]
    \centering
    \includegraphics[width=\linewidth]{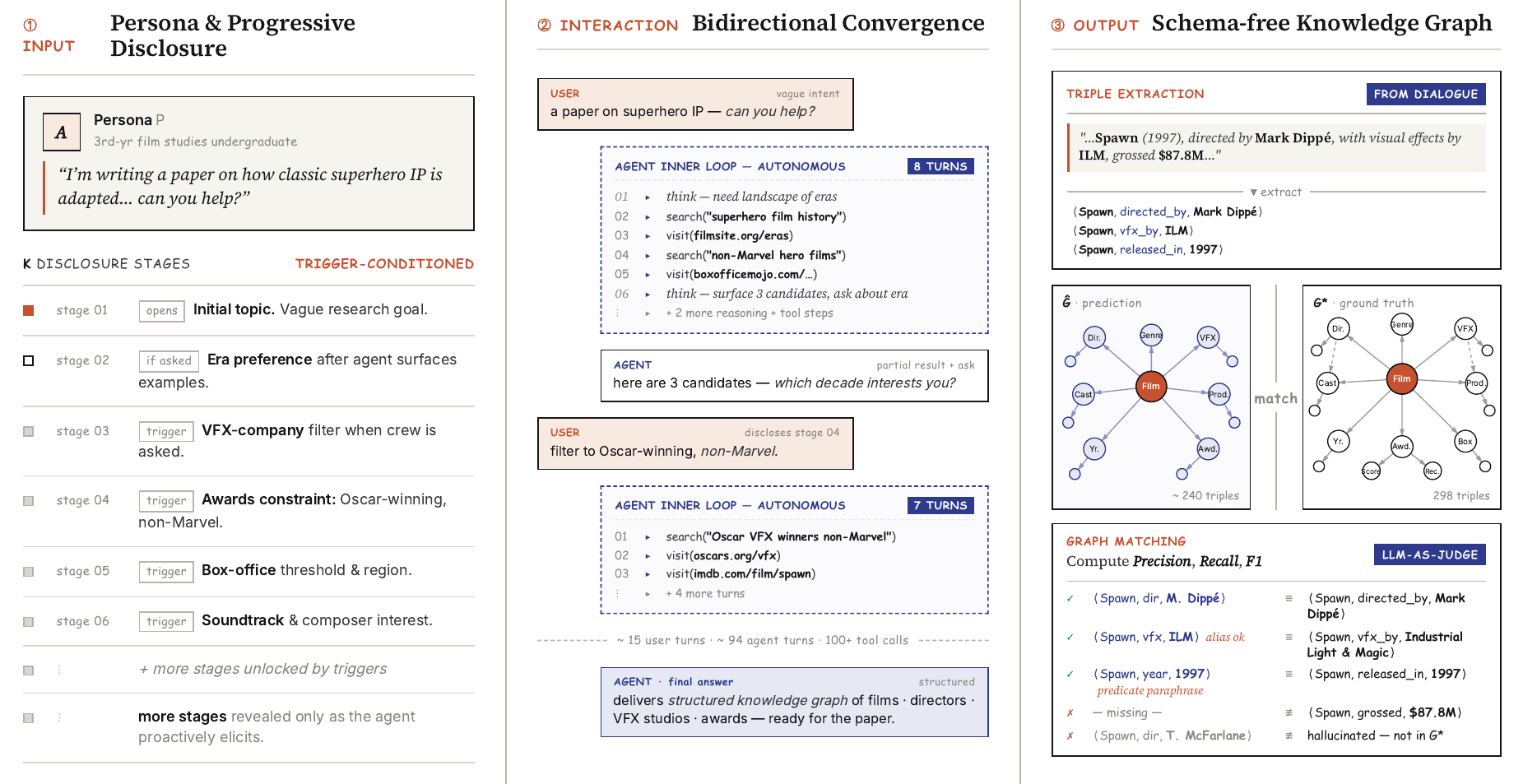}
    \caption{Figure 1: Overview of VibeSearchBench. (Left) A user persona $\mathcal{P}$ with $K$ trigger-conditioned disclosure stages that progressively reveal latent information needs. (Center) A multi-turn bidirectional convergence process in which the agent autonomously executes search and reasoning steps (inner loop), returns partial results, and interacts with user to unlock subsequent stages. (Right) A schema-free knowledge graph output, evaluated by matching the predicted graph $\hat{\mathcal{G}}$ against the ground-truth graph $\mathcal{G}^*$.}
    \label{fig:vibesearchbench}
\end{figure}

Existing mainstream search benchmarks (shown in Table \ref{tab:benchmark_comparison}) fail to capture the VibeSearch paradigm in three critical ways.
\textbf{(1)~Over-specified queries.} Task constraints are exhaustively and explicitly packed into a single prompt (WideSearch, for instance, provides the complete table schema upfront), leaving no room for the agent to actively elicit user intent.
\textbf{(2)~Single-turn interaction.} Current benchmarks do not support sustained user-agent interaction, thereby skipping the most challenging and valuable step in VibeSearch: proactively and continuously mining the user's true search intent.
\textbf{(3)~Fixed-schema outputs and evaluation.} Outputs are evaluated against predetermined structures such as items, sets, or tables. However, real-world knowledge relationships are inherently complex, and user search intent is difficult to model with rigid schemas.

We argue that effective VibeSearch systems should adhere to two principles.
\textbf{First, search should be a process of bidirectional convergence, not unidirectional answering.}
Users often cannot articulate their preferences until they have seen some relevant information; the agent should therefore interleave returning partial results with asking follow-up questions, co-evolving vague needs into concrete solutions with the user, rather than following a ``clarify first, search later'' two-stage pipeline.
 \textbf{Second, outputs and evaluation should be grounded in schema-free structured information}.
Fixed-schema evaluation, while objective and stable, is misaligned with the complex knowledge structures found in the real world \cite{ziomek2026llmwikiracebenchmarkfarllms}; free-text evaluation requires rubric design that is inherently subjective and unstable \cite{sharma2025researchrubricsbenchmarkpromptsrubrics,yang2026onemillionbenchfarlanguageagents,ye2026miroevalbenchmarkingmultimodaldeep}.
We observe that a directed graph without any preset schema can model arbitrary target information relevant to the search intent, while still enabling fine-grained, objectively verifiable evaluation.

To fill this gap, we introduce \textbf{VibeSearchBench}, a benchmark designed to evaluate agents' long-horizon proactive search capabilities.
We manually curate 200 high-quality evaluation tasks spanning two subsets, VibeSearch-Pro (professional scenarios) and VibeSearch-Daily (daily-life scenarios), across 20 domains, with 100 tasks each in Chinese and English.
To ensure distributional diversity, every task covers a distinct topic.
Each task comprises a user persona that specifies the searcher's background and latent intent, together with a ground-truth knowledge graph that encodes the target information in a schema-free directed graph.
Building on these components, we design
(i)~a progressive-disclosure user simulator that incrementally reveals information needs during multi-turn interaction with the agent, and
(ii)~a graph-matching evaluation framework that enables objective and fine-grained assessment of retrieved information.


A benchmark, however, is only as informative as the runtime in which it is evaluated. Today, search is overwhelmingly accessed through agent harnesses~\cite{openclaw,hermes,claudecode} deployed as personal assistants, where users issue vague, evolving queries through multi-turn interaction rather than the fully-specified single-turn prompts assumed by existing benchmarks. By abstracting away precisely this dynamic, current benchmarks cannot tell us how frontier models actually search in deployment—their scores characterize a setting real users will almost never encounter. VibeSearchBench is specifically designed to evaluate frontier models on realistic user search scenarios as they deployed in an agent harness.
We instantiate this evaluation on OpenClaw, a widely adopted production harness, and additionally report ReAct results as a research-side reference baseline. Across seven frontier models, our experiments yield three key findings. First, all models perform poorly: the best model (Claude Opus 4.6) achieves only 30.30 average F1, with higher proactiveness (7-8 tool calls per user turn) correlating with better performance, while excessive resource consumption paradoxically degrades results through context overflow. Second, error analysis reveals three cascading bottlenecks: compressed trajectories suffer 8-12 point F1 drops from information loss, no model successfully reaches the user simulator's completion signal due to inefficient intent elicitation, and models produce structurally flat knowledge graphs that fail to cover the desired knowledge. Third, ablation of three core mechanisms of OpenClaw (sub-agent collaboration, local memory, and life-long memory) shows that none yields significant improvement, indicating that the challenges of VibeSearch demand fundamental model-level advances rather than harness-level architectural enhancements.

\begin{table}[t]
\centering
\caption{Comparison of VibeSearchBench with existing search benchmarks.}
\label{tab:benchmark_comparison}
\resizebox{\linewidth}{!}{%
\begin{tabular}{lcccccc}
\toprule
\textbf{Benchmark} & \textbf{Query Spec.} & \textbf{Interaction} & \textbf{Proactive} & \textbf{User Simulation} & \textbf{Output} & \textbf{Evaluation} \\
\midrule
DeepResearch Bench~\cite{du2026deepresearch} & Full  & Single-turn & \ding{55} & None         & Report                & Rubrics-based       \\
BrowseComp~\cite{wei2025browsecompsimplechallengingbenchmark}         & Full  & Single-turn & \ding{55} & None         & Item                  & Entity Match        \\
WideSearch~\cite{wong2025widesearchbenchmarkingagenticbroad}         & Full  & Single-turn & \ding{55} & None         & Table                 & Table Match      \\
DeepSearchQA~\cite{gupta2026deepsearchqabridgingcomprehensivenessgap} & Full & Single-turn & \ding{55} & None & Item/Set & Entity Match \\
GISA~\cite{zhu2026gisabenchmarkgeneralinformationseeking}               & Full  & Single-turn & \ding{55} & None         & Item/Set/List/Table   & Fixed-Schema Match  \\
InteractComp~\cite{deng2025interactcompevaluatingsearchagents}       & Vague & Multi-turn  & \ding{51} & Simple Rules & Item                  & Entity Match        \\
\textbf{VibeSearchBench}   & Vague & Multi-turn  & \ding{51} & Persona-based Progressive Disclosure & Schema-free Graph & Graph Matching \\
\bottomrule
\end{tabular}%
}
\end{table}

\section{Related Work}

\textbf{Benchmarking Search.}
Existing search benchmarks evaluate agents along the complementary axes of depth and breadth, but largely operate under a fully-specified, single-turn paradigm.
BrowseComp~\cite{wei2025browsecompsimplechallengingbenchmark} and DeepSearchQA~\cite{gupta2026deepsearchqabridgingcomprehensivenessgap} emphasize depth, requiring persistent multi-hop browsing to retrieve hard-to-find facts;
WideSearch~\cite{wong2025widesearchbenchmarkingagenticbroad} instead targets breadth, assessing an agent's ability to aggregate parallel sources into pre-specified tables;
and GISA~\cite{zhu2026gisabenchmarkgeneralinformationseeking} generalizes the output format to items, sets, lists, and tables under fixed-schema matching.
InteractComp~\cite{deng2025interactcompevaluatingsearchagents} introduces ambiguous queries and multi-turn interaction, but its user simulator follows simple rules and outputs are still evaluated as single-entity matches.
In contrast, VibeSearchBench combines persona-driven progressive disclosure with schema-free graph evaluation, jointly capturing the realistic dynamics of evolving intent elicitation and the complex relational structure of real-world information.

\textbf{Benchmarking Agent Harness in the wild.}
As Agent Harnesses rapidly mature into widely-deployed personal-assistant products, a parallel line of work has emerged to benchmark their general agentic capabilities, including Claw-Eval~\cite{ye2026clawevaltrustworthyevaluationautonomous}, ClawBench~\cite{zhang2026clawbenchaiagentscomplete}, WildClawBench~\cite{Ding_WildClawBench}, QwenClawBench~\cite{qwenclawbench1.1}, PinchBench~\cite{pinchbench}, and Claw-Mark~\cite{meng2026clawmarklivingworldbenchmarkmultiturn}.
Notably, the majority of these benchmarks still devote a fraction of their tasks to search- and research-oriented scenarios, reflecting the empirical observation that information acquisition remains one of the most frequent and most demanding user needs once such harnesses are deployed in the wild.
This makes the intersection of agent harnesses and search a particularly consequential setting to study, rather than a niche one.

\section{VibeSearchBench}

\subsection{Task Definition}

We formalize VibeSearch as follows. Each task consists of a user persona $\mathcal{P}$ and a ground-truth knowledge graph $\mathcal{G}^* = (\mathcal{V}^*, \mathcal{E}^*)$, where $\mathcal{V}^*$ is the set of entities and $\mathcal{E}^*$ is the set of triples (each triple $(h, r, t)$ denotes a relation $r$ between a head entity $h$ and a tail entity $t$). $\mathcal{G}^*$ is a schema-free directed graph capable of modeling arbitrary target information relevant to the search intent.

The user persona $\mathcal{P}$ comprises the user's background profile (domain expertise, preferences, etc.), an initial vague query $q_0$, and a sequence of staged information needs $\{(c_k, u_k)\}_{k=1}^{K}$, where $c_k$ is the trigger condition for the $k$-th stage and $u_k$ is the new requirement the user will disclose at that stage.

The search process is modeled as a multi-turn interaction. At turn $t$, the agent takes the dialogue history $\mathcal{H}_t = \{(u_1, a_1), \ldots, (u_{t-1}, a_{t-1})\}$ and available search tools as input, executes search operations, and generates a response $a_t$. The user simulator evaluates whether $a_t$ satisfies the current trigger condition $c_k$: if satisfied, it discloses $u_k$ and advances to the next stage; otherwise, it pushes the agent to continue. The interaction proceeds until all stages are addressed or the budget is exhausted.

After the interaction concludes, the agent organizes all gathered information into a predicted knowledge graph $\hat{\mathcal{G}} = (\hat{\mathcal{V}}, \hat{\mathcal{E}})$, output as a list of triples. Evaluation computes triplet-level precision, recall, and F1 via graph matching between $\hat{\mathcal{G}}$ and $\mathcal{G}^*$.

\subsection{Construction Pipeline}
\textbf{Expert Annotation.}
We recruit professional annotators from 20 domains. Each annotator is required to: (1)~design a plausible search scenario with an initial vague query $q_0$; (2)~simulate a multi-turn interaction with an AI assistant, progressively refining their search needs; and (3)~construct a ground-truth knowledge graph $\mathcal{G}^*$ whose nodes and triples are consistent with the search intent and the information ultimately obtained. To ensure distributional diversity, every task covers a distinct topic. This process yields 200 tasks spanning VibeSearch-Pro (professional domains) and VibeSearch-Daily (everyday scenarios), with 100 tasks each in Chinese and English.

\textbf{User Persona Synthesis.}
Based on the annotated multi-turn queries and ground-truth graphs, we synthesize structured user personas $\mathcal{P}$. Each persona defines $K$ information-disclosure stages, where each stage specifies: (1)~a trigger condition $c_k$ (e.g., the agent proactively asks about a certain aspect, or the response contains specific information); (2)~the user's response content $u_k$ when the condition is met; and (3)~behavioral strategies when the condition is not met (e.g., pushing the agent to continue, commenting on results, or requesting more details). The original annotators review and revise each persona to ensure consistency with $\mathcal{G}^*$.

\textbf{Quality Control.}
We adopt a dual-review mechanism to ensure data quality. After each task is annotated, it is independently reviewed by two domain experts who are not among the annotators. The review covers: (1) the rationality and authenticity of the search scenario; (2) the naturalness and logical coherence of the multi-turn interaction flow; (3) whether the progressive disclosure of information needs is reasonable; (4) the correctness of factual information in the ground truth graph; and (5) the consistency between the user persona and the ground truth graph. Both reviewers' opinions must be approved simultaneously; any task that fails on any dimension will be returned to the annotator for revision or redoing until all quality criteria are met.

\begin{wraptable}{r}{0.5\textwidth}
\centering
\small
\vspace{-0.7cm}
\caption{Statistics of VibeSearchBench.}
\label{tab:statistics}
\resizebox{0.5\textwidth}{!}{%
\begin{tabular}{lccc}
\toprule
\textbf{Statistic} & \textbf{Pro} & \textbf{Daily} & \textbf{Total} \\
\midrule
\# Tasks & 100 & 100 & 200 \\
\# Chinese tasks & 50 & 50 & 100 \\
\# English tasks & 50 & 50 & 100 \\
\# Domains & 5 & 15 & 20 \\
Avg. \# GT nodes & 278.68 & 146.18 & 212.43 \\
Avg. \# GT triples & 373.56 & 223.07 & 298.32 \\
Avg. \# source URLs & 158.29 & 121.11 & 139.70 \\
Avg. URLs / triples & 0.42 & 0.54 & 0.47 \\
\bottomrule
\end{tabular}%
}
\end{wraptable}

\subsection{User Simulator}

The user simulator drives multi-turn interactions by taking the persona $\mathcal{P}$ and the agent's response $a_t$ to generate the user's reply. It follows four core principles: (1)~Progressive disclosure: information needs are disclosed one stage at a time, forcing the agent to proactively unlock deeper needs. (2)~Condition-driven transitions: each stage advances only when an explicit trigger condition is met (e.g., the agent mentions specific information, asks about a relevant aspect, or completes a milestone). (3)~Persistent pressure: when conditions are unmet, the simulator continues engaging by commenting on results, requesting details, or urging completion. (4)~Natural conversation: the simulator responds to every agent question, including irrelevant ones (e.g., ``no particular preference''), ensuring interaction realism. We use an LLM as the backbone, encoding these principles into behavioral rules via a system prompt. We show the prompt in \ref{tab:user_simulator_prompt}.

\begin{wrapfigure}{r}{0.5\textwidth}
\centering
\vspace{-10pt}
\includegraphics[width=0.5\textwidth]{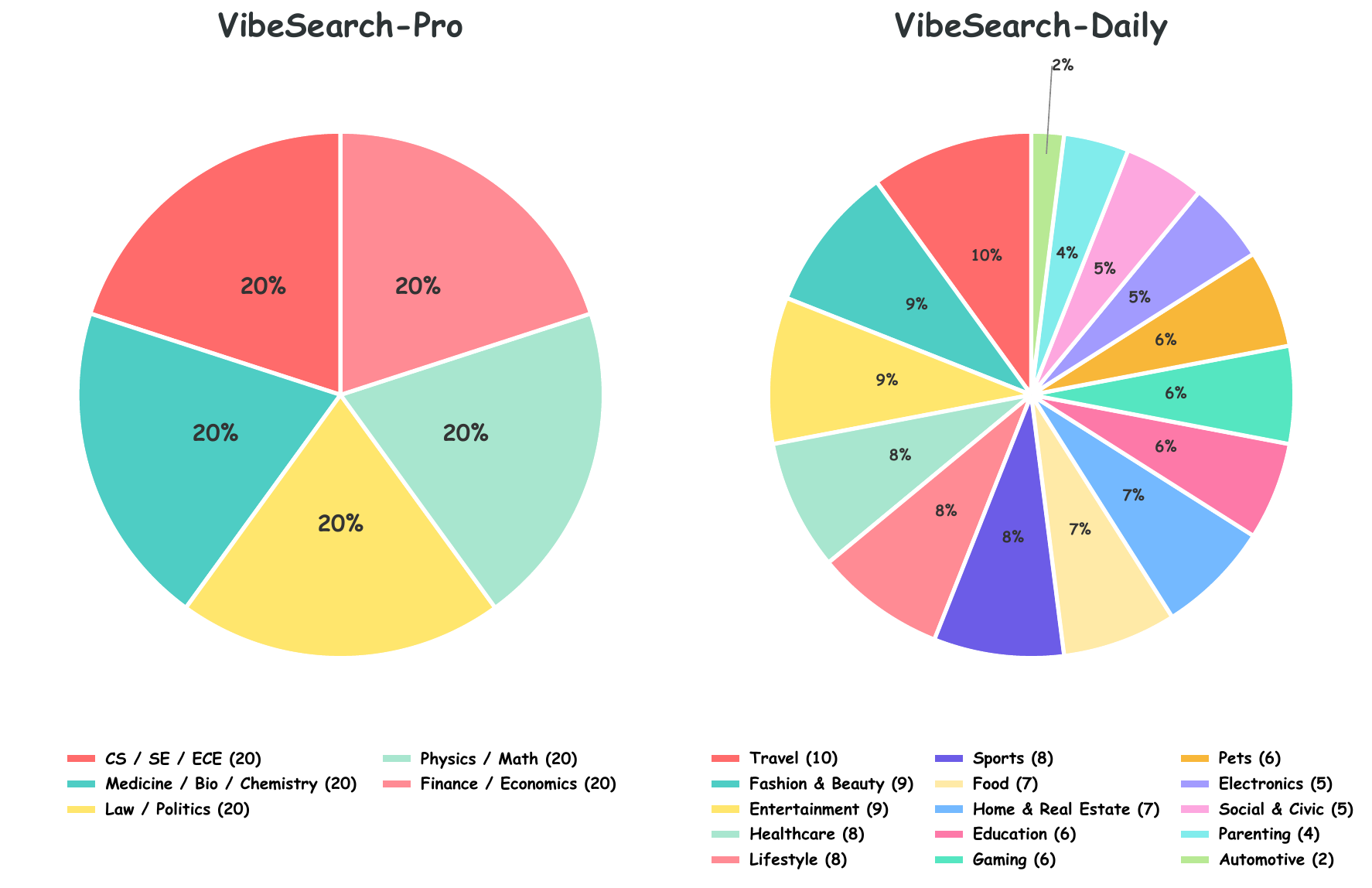}
\caption{Domain distribution of VibeSearchBench.}
\label{fig:domain_distribution}
\vspace{-10pt}
\end{wrapfigure}

\subsection{Graph-based Evaluation}
\label{sec:evaluation}
We propose an information-entailment-based evaluation framework that uses an LLM-as-judge to perform graph matching, accommodating semantically equivalent expressions (e.g., entity aliases, relation synonyms) unlike exact matching.
For \textbf{recall}, the judge determines whether each ground-truth triple is ``covered'' by the predicted graph, considering direct matches, subsumption, collective coverage by multiple triples, or compositional derivation through existing predicted relations. \textbf{Precision} is computed as the fraction of predicted triples that participate in covering at least one ground-truth triple. \textbf{F1} is the harmonic mean of precision and recall. Ground-truth triples are partitioned into batches and evaluated in parallel for efficiency. Formal details are provided in Appendix~\ref{sec:eval_details}.

\subsection{Statistics}


Table~\ref{tab:statistics} presents the overall statistics of VibeSearchBench. The benchmark contains 200 tasks, evenly split into VibeSearch-Pro (professional domains) and VibeSearch-Daily (daily life), with 100 Chinese and 100 English tasks covering 20 distinct domains.
Each task's ground truth graph contains 212.43 nodes and 298.32 triples on average, reflecting the richness of information required. VibeSearch-Pro graphs are notably larger than VibeSearch-Daily ones (373.56 vs. 223.07 triples), indicating that professional-domain tasks involve more complex knowledge structures.
Each task involves 139.70 distinct source URLs on average, with a URL-to-triple ratio of 0.47, indicating that multiple facts are typically extracted per source. This ratio is higher for VibeSearch-Daily (0.54) than VibeSearch-Pro (0.42), suggesting that daily-life information sources are more dispersed and individually less informative.
The representative examples are provided in the appendix \ref{app:examples}.

\section{Experiments}

\subsection{Experimental Setting}
\label{sec:experiment}
\textbf{Models.}
We evaluate seven frontier LLMs on VibeSearchBench: Claude Opus 4.6 \cite{opus4_6}, GPT-5.4 \cite{gpt5_4}, Gemini-3.1 Pro \cite{gemini3_1}, Seed2.0 Pro \cite{seed2}, Kimi K2.6 \cite{kimi2_6}, DeepSeek-V4-Pro \cite{deepseek_v4} , and Qwen-3.5-397B-A17B \cite{qwen3_5}. These models cover both proprietary and open-source frontier models.

\textbf{Agent Frameworks.}
We conduct under: (1)~ReAct, the classic reasoning-and-acting framework in which the agent alternates between reasoning and tool execution at each step; and (2)~OpenClaw, a rapidly maturing agent harness that is widely adopted as a personal assistant. Comparing the two frameworks aims to reveal how different interaction paradigms affect VibeSearch performance.

\textbf{Implementation Details.}
All models are run with default parameters; the search tool configuration is detailed in Appendix~\ref{app:tools}. We set the max context window as 256k. For ReAct, we equip it with a simple compaction mechanism to handle context-overflow situations: when the model's context is about to exceed 256k tokens, we have it summarize its own context and then continue interacting with the user based on this summary. We use Seed-2.0-Pro as the backbone model for the user simulator. Each model is run 3 times per task, and we report the averaged result. We adopt the triplet-level Precision, Recall, and F1 defined in Section~\ref{sec:evaluation} as the evaluation metrics.

\subsection{Main Results}

Table~\ref{tab:main_results} presents the results of all models under both frameworks.

\textbf{Overall.}
Even the strongest model, Claude Opus 4.6, achieves only 30.30 average F1 under OpenClaw, and all models score below 33, indicating that current models remain substantially inadequate for VibeSearch.
A clear hierarchy emerges: Claude Opus 4.6 and DeepSeek-V4-Pro form the top tier (F1 $\geq$27), followed by Kimi K2.6 in the middle range, with GPT-5.4 and Qwen3.5-397B-A17B trailing (20--23).
OpenClaw slightly outperforms ReAct on most models (Claude +2.43, GPT +1.88), but Kimi K2.6 (26.09 vs.\ 26.17) and Gemini-3.1 Pro (23.54 vs.\ 23.62) show no meaningful difference, suggesting that the benefit of an agent harness depends on the underlying model's capability. Seed2.0 Pro's Daily F1 improves notably under OpenClaw (20.58 $\to$ 24.64), indicating that weaker models may benefit more from framework support.

\textbf{Precision vs.\ Recall.}
Most models exhibit Recall $>$ Precision (e.g., Claude: P=24.88, R=36.34), favoring broad coverage at the cost of many irrelevant triples. This imbalance is especially pronounced on Daily, where Claude's Recall reaches 39.20 while Precision drops to 21.60.
The sole exception is Gemini-3.1 Pro (P=34.61, R=20.63), which conservatively outputs high-confidence information but leaves nearly 84\% of ground-truth triples on Pro unrecovered.
Kimi K2.6 achieves the most balanced profile (P=28.29, R=27.52), avoiding both over-generation and under-exploration.

\textbf{Pro vs.\ Daily.}
Pro subset F1 is consistently higher than Daily (e.g., Claude: 29.79 vs.\ 25.95; DeepSeek: 28.70 vs.\ 25.37), as professional domains feature concentrated, well-structured information. Daily scenarios are harder because (1)~information is more scattered (URL-to-triple ratio 0.54 vs.\ 0.42) and (2)~user needs are more diverse and harder to anticipate.
Gemini-3.1 Pro is a notable exception, achieving higher F1 on Daily (24.66 vs.\ 22.41), because its snippet-only strategy is less penalized when ground-truth graphs are smaller (Daily: 223 triples vs.\ Pro: 374).

\begin{table}[t]
\centering
\caption{Performance of all models on VibeSearchBench. The upper section shows results under the ReAct framework; the lower section shows results under the OpenClaw framework.}
\label{tab:main_results}
\resizebox{\linewidth}{!}{%
\begin{tabular}{l ccc ccc ccc}
\toprule
& \multicolumn{3}{c}{\textbf{VibeSearch-Pro}} & \multicolumn{3}{c}{\textbf{VibeSearch-Daily}} & \multicolumn{3}{c}{\textbf{Average}} \\
\cmidrule(lr){2-4} \cmidrule(lr){5-7} \cmidrule(lr){8-10}
\textbf{Model} & P & R & F1 & P & R & F1 & P & R & F1 \\
\midrule
\multicolumn{10}{l}{\textit{ReAct}} \\
\midrule
Claude Opus 4.6    & 28.15 & \textbf{33.47} & \textbf{29.79} & 21.60 & \textbf{39.20} & \textbf{25.95} & 24.88 & \textbf{36.34} & \textbf{27.87} \\
DeepSeek-V4-Pro    & 29.81 & 29.54 & 28.70 & 21.81 & 35.41 & 25.37 & 25.81 & 32.48 & 27.04 \\
Gemini-3.1 Pro     & \textbf{40.88} & 16.00 & 22.41 & \textbf{28.33} & 25.25 & 24.66 & \textbf{34.61} & 20.63 & 23.54 \\
GPT-5.4            & 19.54 & 19.42 & 18.94 & 18.02 & 30.06 & 21.13 & 18.78 & 24.74 & 20.04 \\
Kimi K2.6          & 33.11 & 23.80 & 27.12 & 23.47 & 31.24 & 25.05 & 28.29 & 27.52 & 26.09 \\
Qwen3.5-397B-A17B  & 23.90 & 25.40 & 23.65 & 20.09 & 29.19 & 22.02 & 22.00 & 27.30 & 22.84 \\
Seed2.0 Pro        & 30.25 & 23.61 & 25.86 & 18.16 & 28.95 & 20.58 & 24.21 & 26.28 & 23.22 \\
\midrule
\multicolumn{10}{l}{\textit{OpenClaw}} \\
\midrule
Claude Opus 4.6    & 29.24 & \textbf{38.13} & \textbf{32.55} & 24.51 & \textbf{35.90} & \textbf{28.04} & 26.88 & \textbf{37.02} & \textbf{30.30} \\
DeepSeek-V4-Pro    & 28.68 & 31.43 & 29.11 & 22.13 & 35.13 & 26.16 & 25.41 & 33.28 & 27.64 \\
Gemini-3.1 Pro     & \textbf{39.48} & 15.48 & 21.69 & \textbf{29.37} & 24.95 & 25.54 & \textbf{34.43} & 20.22 & 23.62 \\
GPT-5.4            & 23.52 & 22.99 & 22.78 & 19.22 & 25.64 & 21.05 & 21.37 & 24.32 & 21.92 \\
Kimi K2.6          & 32.41 & 24.80 & 27.40 & 23.99 & 28.32 & 24.93 & 28.20 & 26.56 & 26.17 \\
Qwen3.5-397B-A17B  & 26.19 & 24.03 & 24.42 & 19.80 & 26.72 & 21.77 & 23.00 & 25.38 & 23.10 \\
Seed2.0 Pro        & 33.22 & 22.12 & 25.81 & 23.73 & 27.53 & 24.64 & 28.48 & 24.83 & 25.23 \\
\bottomrule
\end{tabular}%
}
\end{table}

\subsection{Interaction Behavior}

\begin{table}[ht]
\centering
\caption{Interaction behavior statistics on VibeSearchBench (averaged over Pro and Daily). \# Asst and \# User denote the number of agent and user dialogue turns, respectively. \#Asst/\#User reflects the agent's average work intensity per user turn. \# Compact denotes the average number of context compressions.}
\label{tab:interaction}
\resizebox{\linewidth}{!}{%
\begin{tabular}{l cccc cccc}
\toprule
& \multicolumn{4}{c}{\textit{ReAct}} & \multicolumn{4}{c}{\textit{OpenClaw}} \\
\cmidrule(lr){2-5} \cmidrule(lr){6-9}
\textbf{Model} & \# Asst & \# User & \#Asst/\#User & \# Compact & \# Asst & \# User & \#Asst/\#User & \# Compact \\
\midrule
Claude Opus 4.6    & \textbf{109.8} & 13.3 & \textbf{8.26} & 0.68 & \textbf{93.6} & 12.8 & \textbf{7.31} & 0.46 \\
DeepSeek-V4-Pro    & 74.0 & 13.9 & 5.30 & 0.16 & 82.6 & 15.7 & 5.28 & 0.29 \\
Gemini-3.1 Pro     & 41.0 & 14.4 & 2.84 & 0.00 & 51.2 & 14.9 & 3.43 & 0.01 \\
GPT-5.4            & 99.6 & \textbf{15.3} & 6.51 & \textbf{1.27} & 86.5 & \textbf{19.9} & 4.34 & \textbf{1.36} \\
Kimi K2.6          & 72.0 & 15.1 & 4.77 & 0.07 & 55.4 & 14.8 & 3.73 & 0.08 \\
Qwen3.5-397B-A17B  & 67.2 & 13.6 & 4.96 & 0.17 & 54.7 & 13.9 & 3.93 & 0.15 \\
Seed2.0 Pro        & 73.0 & 14.9 & 4.88 & 0.01 & 84.8 & 15.9 & 5.31 & 0.09 \\
\bottomrule
\end{tabular}%
}
\end{table}

\begin{figure}[ht]
\centering
\includegraphics[width=\linewidth]{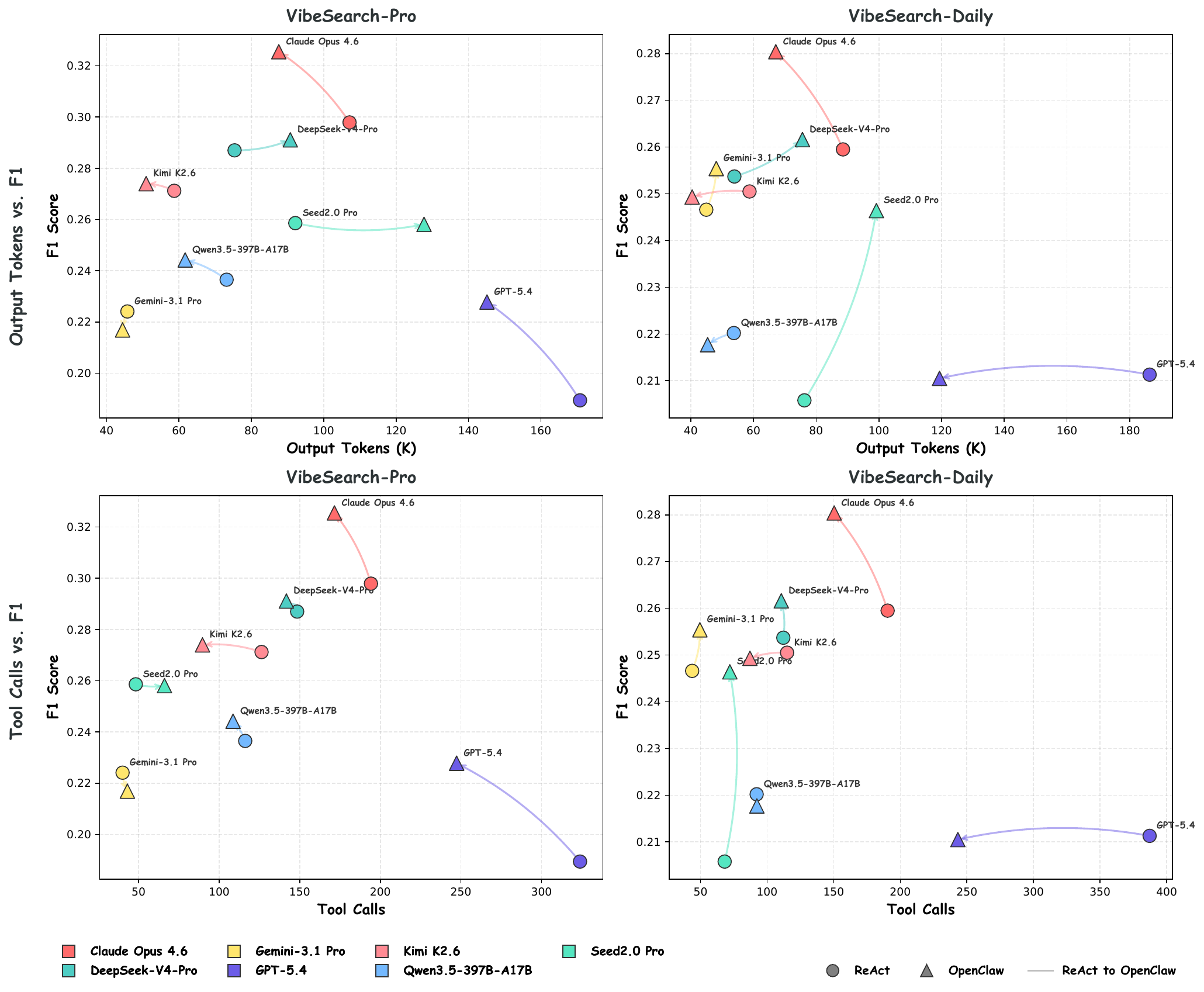}
\caption{Resource consumption vs.\ F1 score. Top row: output tokens vs.\ F1; bottom row: total tool calls vs.\ F1. Each model appears twice (circle for ReAct, triangle for OpenClaw), with an arrow indicating the shift when switching frameworks.}
\label{fig:cost_vs_f1}
\end{figure}

Table~\ref{tab:interaction} presents the interaction behavior statistics of all models under both frameworks.

\textbf{Proactiveness.}
The \#Asst/\#User ratio measures the amount of independent search and reasoning work the agent performs between user turns; a higher ratio indicates stronger proactiveness.
Claude Opus 4.6 achieves the highest ratio (ReAct: 8.26), executing 7--8 tool calls per user reply on average, and also the highest F1, demonstrating a direct link between proactiveness and performance.
Gemini-3.1 Pro has the lowest ratio (2.84), passively waiting for user-driven exploration, resulting in severely limited coverage.

\textbf{Interaction Efficiency.}
Claude Opus 4.6 has the fewest user turns (ReAct: 13.3), advancing information disclosure most efficiently.
GPT-5.4 is a notable counter-example: despite high assistant turns (99.6), its user turns are also the highest (OpenClaw: 19.9), yielding an unremarkable \#Asst/\#User ratio (4.34).
More critically, its context compression count far exceeds all other models (1.27 vs.\ $<$0.7 for others), as verbose output triggers frequent context overflow that destroys previously retrieved information and forces redundant re-searching, creating a vicious cycle of ``verbose output $\to$ context overflow $\to$ information loss $\to$ performance degradation'' that fundamentally explains its worst F1 despite the highest resource consumption.

\textbf{Framework Effects on Interaction Patterns.}
Claude's assistant turns decrease under OpenClaw (109.8 $\to$ 93.6) while F1 improves (27.87 $\to$ 30.30), indicating higher efficiency per turn.
Seed2.0 Pro shows the opposite pattern: assistant turns increase (73.0 $\to$ 84.8) alongside F1 improvement (23.22 $\to$ 25.23), benefiting from the expanded exploration space.

\subsection{Cost-Performance}

\textbf{No Positive Correlation Between Resource Consumption and Performance.} Figure~\ref{fig:cost_vs_f1} shows the relationship between each model's output token count and tool call count versus F1.
As shown, resource consumption is not positively correlated with F1.
GPT-5.4 consumes the most resources (both output tokens and tool calls far exceed other models) yet scores the lowest F1, as verbose output triggers frequent context compression that reduces subsequent searches to redundant work.
Gemini-3.1 Pro has the lowest resource consumption and almost never uses the visit tool (Pro: 0.05 times), resulting in severely insufficient information acquisition depth.
Claude Opus 4.6 and DeepSeek-V4-Pro achieve the best F1 at moderate resource levels, suggesting an efficiency sweet spot: too little exploration limits coverage, while excessive exploration degrades performance through context management burden.

\section{Analysis}

\subsection{Error Analysis}
\label{sec:error_analysis}

We analyze all ReAct trajectories and categorize failures along three pipeline stages (Table~\ref{tab:error_analysis}). These failures cascade: context overflow during retrieval causes agents to forget previously disclosed requirements, producing misaligned output downstream. The complete error analysis is shown in Appendix \ref{app:error_analysis}.

\textbf{Information Retrieval and Context Management Failures.}
Models are trapped between two symmetric failures: context overflow from excessive exploration versus information gaps from conservative retrieval.
As shown in the Comp.\% and $\Delta$F1 columns of Table~\ref{tab:error_analysis}, compressed trajectories suffer a consistent 8--12 point F1 drop (0.16 vs.\ 0.26 on average).
GPT-5.4 exemplifies the former: with the highest compression rate (72.0\%), its F1 declines from 0.25 with zero compressions to 0.12 with two or more (Table~\ref{tab:compression_gpt}), as verbose output triggers a compounding overflow cycle.
Gemini-3.1 Pro exemplifies the latter: it avoids compression entirely (0.0\%) but almost never visits pages beyond search snippets (averaging only 1.1 page visits per task on Pro); on Daily, trajectories where Gemini visits at least one page achieve 55\% higher Recall (0.34 vs.\ 0.22; Appendix~\ref{app:error_analysis}).
Kimi K2.6 strikes the best balance with moderate search volume and the lowest compression rate among actively searching models (6.8\%).

\textbf{Multi-Turn Interaction and Intent Elicitation Failures.}
Virtually no trajectory across all runs reaches the user simulator's \texttt{[DONE]} signal; all terminate via agent-initiated answer or \texttt{max\_rounds} exhaustion.
As detailed in Table~\ref{tab:turn_f1_corr}, trajectories exceeding 15 user turns average only 0.18 F1 versus 0.23 for those with $\leq$10, reflecting both intrinsically harder tasks (requiring more rounds due to scattered information) and wasted turns on misaligned questions.
We further measure the fraction of user messages containing dismissive patterns (indicating failed intent elicitation) and redirect patterns (indicating premature stage advancement; Table~\ref{tab:interaction_strategy}).
Gemini-3.1 Pro's passive strategy (\#Asst/\#User ratio of 2.84) yields the highest dismissive response rate (7.9\% on Daily), while redirect rates remain uniformly at 3--6\% across all models, revealing a universal tendency to advance stages before fully satisfying current requirements.

\textbf{Knowledge Graph Construction and Output Failures.}
We analyze the structural alignment between predicted and ground-truth knowledge graphs by examining per-relation coverage rates (Table~\ref{tab:relation_coverage}).
Models extract facts effectively but fail to organize them hierarchically: even the best model achieves 100\% coverage on factual relations (e.g., \texttt{participating\_country}, \texttt{restructuring\_year}) yet 0\% on organizational and hierarchical ones (e.g., \texttt{includes\_phase}, \texttt{case\_participated}), producing only flat, instance-level triples.
As reflected in the Over\% and Under\% columns of Table~\ref{tab:error_analysis}, this structural gap yields two divergent failure modes.
\emph{Over-generation} affects 54\% of Claude trajectories, primarily driven by bibliographic metadata extraction (99\% invalidity rate) and subjective assessments (95--99\% invalidity rate; Table~\ref{tab:invalid_predictions}).
\emph{Under-generation} dominates 46\% of Gemini trajectories due to conservative retrieval that leaves most information unextracted.
Format failures further cause catastrophic collapse: Seed2.0 Pro produces 28 zero-F1 trajectories (the most among all models) from malformed JSON output, underscoring that knowledge graph construction remains fragile under long interaction histories.

\begin{table}[t]
\centering
\vspace{-1em}
\caption{Error analysis under the ReAct framework (averaged over Pro and Daily). Comp.\% = fraction of trajectories that trigger context compression; F1(C) and F1(NC) = mean Triplet F1 for compressed and non-compressed trajectories; $\Delta$F1 = F1(C)$-$F1(NC); 0-F1 = number of zero-F1 trajectories; Over\% and Under\% = fraction of trajectories exhibiting over-generation and under-generation, respectively. Bold indicates the most extreme value in each column. Detailed breakdowns are in Appendix~\ref{app:error_analysis}.}
\label{tab:error_analysis}
\resizebox{\linewidth}{!}{%
\begin{tabular}{lccccccc}
\toprule
\textbf{Model} & \textbf{Comp.\%} & \textbf{F1(C)} & \textbf{F1(NC)} & $\boldsymbol{\Delta}$\textbf{F1} & \textbf{0-F1} & \textbf{Over\%} & \textbf{Under\%} \\
\midrule
Claude Opus 4.6    & 62.0 & \textbf{0.183} & \textbf{0.275} & $-$0.092 & 6  & \textbf{54.0} & 0.8  \\
GPT-5.4            & \textbf{72.0} & 0.175 & 0.259 & $\boldsymbol{-}$\textbf{0.084} & 3  & 28.7 & 2.0  \\
DeepSeek-V4-Pro    & 16.0 & 0.150 & 0.256 & $-$0.106 & 4  & 35.2 & 3.4  \\
Kimi K2.6          & 6.8  & 0.151 & 0.273 & $-$0.122 & 5  & 12.3 & 8.7  \\
Qwen3.5-397B-A17B  & 16.0 & 0.135 & 0.247 & $-$0.112 & 7  & 24.7 & 4.5  \\
Seed2.0 Pro        & 0.8  & ---   & 0.205 & ---      & \textbf{28} & 28.7 & 4.7  \\
Gemini-3.1 Pro     & 0.0  & ---   & 0.231 & ---      & 6  & 4.2  & \textbf{46.0} \\
\bottomrule
\end{tabular}%
}
\end{table}

\subsection{OpenClaw Analysis}
\label{sec:openclaw_analysis}

We use Kimi K2.6 and Qwen3.5-397B-A17B as base models to ablate three core mechanisms of the OpenClaw framework: sub-agent collaboration, local memory, and life-long memory. All experiments are repeated 3 times and averaged. Auxiliary metrics (\#Asst, \#Tools, Compact) are from the Pro subset.

\begin{wraptable}{r}{0.5\textwidth}
\centering
\vspace{-0.3cm}
\caption{Sub-agent ablation. For +Sub-agent, \#Asst and \#Tools are the sum of the main agent and all sub-agents.}
\label{tab:subagent}
\resizebox{\linewidth}{!}{%
\begin{tabular}{llcccccc}
\toprule
\textbf{Model} & \textbf{Setting} & \textbf{F1 P} & \textbf{F1 D} & \textbf{\#Sub} & \textbf{\#Asst} & \textbf{\#Tools} & \textbf{Comp.} \\
\midrule
Kimi K2.6 & Naive      & 27.40 & 24.93 & 0.0 & 59.9  & 89.7  & 0.11 \\
Kimi K2.6 & +Sub-agent & 26.45 & 23.47 & 4.0 & 98.2  & 163.1 & 0.13 \\
Qwen3.5   & Naive      & 24.42 & 21.77 & 0.0 & 61.1  & 108.6 & 0.20 \\
Qwen3.5   & +Sub-agent & 25.95 & 22.33 & 8.2 & 140.0 & 248.4 & 0.08 \\
\bottomrule
\end{tabular}%
}
\end{wraptable}

\textbf{Sub-agent.}
Delegating retrieval to 4.0--8.2 child agents substantially increases workload (\#Asst +64\%--129\%, \#Tools +82\%--129\%), yet F1 shows no consistent improvement (Kimi Pro $-$0.95, Qwen Pro +1.53).
Qwen's compression drops from 0.20 to 0.08, confirming that sub-agents offload context pressure, but F1 barely improves because cross-agent information coordination incurs significant loss during integration by the main agent.

\begin{wraptable}{r}{0.5\textwidth}
\centering
\vspace{-0.3cm}
\caption{Local memory ablation. Mem = avg memory ops per task. Adopt.\ = fraction of tasks using memory.}
\label{tab:local_memory}
\resizebox{\linewidth}{!}{%
\begin{tabular}{llccccccc}
\toprule
\textbf{Model} & \textbf{Setting} & \textbf{F1 P} & \textbf{F1 D} & \textbf{\#Asst} & \textbf{\#Tools} & \textbf{Comp.} & \textbf{Mem} & \textbf{Adopt.} \\
\midrule
Kimi K2.6 & Naive  & 27.40 & 24.93 & 59.9 & 89.7  & 0.11 & 0.01 & 1.0\%  \\
Kimi K2.6 & +Local & 26.94 & 24.48 & 74.9 & 101.9 & 0.23 & 7.99 & 76.3\% \\
Qwen3.5   & Naive  & 24.42 & 21.77 & 61.1 & 108.6 & 0.20 & 0.00 & 0.3\%  \\
Qwen3.5   & +Local & 24.44 & 21.27 & 63.1 & 90.3  & 0.25 & 2.22 & 53.2\% \\
\bottomrule
\end{tabular}%
}
\end{wraptable}

\textbf{Local Memory.}
Despite high adoption (Kimi 76.3\%, Qwen 53.2\%), F1 remains unchanged ($\pm$0.5).
Memory operations impose significant context overhead: Kimi's assistant turns increase by 25\% and compressions double (0.11$\to$0.23); Qwen redirects effort from retrieval to memory maintenance (\#Tools $-$17\%), yet compressions still rise. The persistence benefits are offset by the context pressure introduced.

\textbf{Life-long Memory.}
F1 differences across all conditions remain below 1.0, indicating that cross-task knowledge transfer fails to take effect.
Life-long memory barely alters behavior: Kimi's \#Asst and \#Tools are identical to naive (52.3 vs.\ 52.5, 76.8 vs.\ 76.8), yet compressions still increase (0.07$\to$0.14, Qwen 0.18$\to$0.27).
The two models exhibit distinct failure modes: Kimi actively queries pre-built memory (84.0\% adoption) but retrieved strategies are too generic; Qwen largely ignores it (19.3\% adoption), reverting to standard behavior.

\begin{wraptable}{r}{0.5\textwidth}
\centering
\vspace{0cm}
\caption{Life-long memory ablation (last 50 tasks; first 50 build the memory store). Mem and Adopt.\ defined as in Table~\ref{tab:local_memory}.}
\label{tab:lifelong_memory}
\resizebox{\linewidth}{!}{%
\begin{tabular}{llccccccc}
\toprule
\textbf{Model} & \textbf{Setting} & \textbf{F1 P} & \textbf{F1 D} & \textbf{\#Asst} & \textbf{\#Tools} & \textbf{Comp.} & \textbf{Mem} & \textbf{Adopt.} \\
\midrule
Kimi K2.6 & Naive      & 27.60 & 22.64 & 52.5 & 76.8 & 0.07 & 0.01 & 1.3\%  \\
Kimi K2.6 & +Local     & 27.60 & 22.42 & 68.7 & 86.9 & 0.15 & 8.91 & 78.7\% \\
Kimi K2.6 & +Life-long & 27.77 & 22.23 & 52.3 & 76.8 & 0.14 & 1.68 & 84.0\% \\
Qwen3.5   & Naive      & 24.85 & 19.83 & 59.5 & 98.0 & 0.18 & 0.00 & 0.0\%  \\
Qwen3.5   & +Local     & 25.70 & 18.57 & 62.5 & 80.9 & 0.21 & 2.15 & 54.7\% \\
Qwen3.5   & +Life-long & 24.85 & 19.18 & 54.7 & 86.2 & 0.27 & 0.25 & 19.3\% \\
\bottomrule
\end{tabular}%
}
\end{wraptable}

\textbf{Summary.}
All three core mechanisms of frontier agent harnesses fail to significantly improve VibeSearch performance, revealing that the required capabilities (evolving intent understanding, scattered information integration, and structured knowledge construction) cannot be solved through external architectural enhancements alone. The key lies in fundamental model advances: stronger long-context integration and more precise intent modeling.

\subsection{Meta-Evaluation Analysis}
\label{sec:meta_eval}

\begin{wraptable}{r}{0.5\textwidth}
\centering
\vspace{-0.5cm}
\caption{Meta-evaluation: agreement between LLM judges and human on ground-truth triple recall.}
\label{tab:meta_eval}
\resizebox{\linewidth}{!}{%
\begin{tabular}{lccc}
\toprule
\textbf{Judge Model} & \textbf{Overall} & \textbf{Covered} & \textbf{Not Covered} \\
\midrule
Qwen3.5-397B-A17B & 98.76\% & 99.36\% & 98.16\% \\
Kimi K2.6         & 98.92\% & 98.64\% & 99.20\% \\
Seed2.0 Pro       & 98.56\% & 98.00\% & 99.12\% \\
\bottomrule
\end{tabular}%
}
\end{wraptable}

We randomly sample 50 trajectories for domain experts to review, and use three LLM judges (Qwen3.5-397B-A17B, Kimi K2.6, Seed2.0 Pro). All three achieve overall agreement above 98.5\% with human experts (Kimi highest at 98.92\%), confirming that the evaluation framework reliably substitutes for human annotation.

\section{Conclusion}

We introduced VibeSearchBench, a benchmark for evaluating LLM agents on long-horizon proactive search, where agents must collaboratively refine vague user intent through multi-turn interaction and produce schema-free information graphs.
Evaluation of seven frontier models under both ReAct and OpenClaw shows that even the best model achieves only 30.30 F1, with context overflow, inefficient intent elicitation, and structurally flat knowledge graph outputs identified as key bottlenecks. Ablation further confirms that architectural enhancements (sub-agents, local memory, life-long memory) yield no meaningful gains. Moreover, the inconsistent framework effects across models (e.g., OpenClaw improves Claude but leaves Kimi unchanged) underscore that optimizing for widely adopted agent harnesses is critical for real-world deployment. 


\section*{Contribution}

\noindent
Z.Y.$^{1,\dagger}$, S.L.$^{1,\dagger}$, Lei Huang$^{1,\dagger}$, Tao Jiang$^{1,\dagger}$, Yunfan Zhang$^{2}$, Jiajie Wu$^{2}$, Yida Zhao$^{2}$, Jialong Wu$^{2}$, Kuan Li$^{2}$, Suyang Wu$^{2}$, XingYu$^{1}$, Xiang Cheng$^{1,\ddagger}$

\vspace{4pt}
\noindent
{\small \footnote{\url{https://studio.dots.ai}}General Post-training Team, Xiaohongshu Dots Studio}\\
{\small \footnote{\url{https://unipat.ai/}}UniPat AI}\\
{\small $^{\dagger}$\,Core Contributor \quad $^{\ddagger}$\,Project Lead}

\newpage
\bibliographystyle{unsrtnat}
\bibliography{references}

\appendix
\newpage
\section{Evaluation Details}
\label{sec:eval_details}

This appendix provides the formal definitions and implementation details of the graph-based evaluation framework described in Section~\ref{sec:evaluation}.

\subsection{Triplet Recall}

For each triple $e^* \in \mathcal{E}^*$ in the ground-truth graph $\mathcal{G}^*$, we use an LLM-as-judge to determine whether the predicted graph $\hat{\mathcal{G}}$ entails the factual information expressed by that triple. A ground-truth triple is judged as ``covered'' if and only if any of the following conditions holds:
\begin{enumerate}[nosep]
    \item A predicted triple directly expresses the same information;
    \item A predicted triple carries more information and subsumes the ground-truth triple;
    \item Multiple predicted triples collectively cover the ground-truth triple's information;
    \item Multiple predicted triples can be composed through explicit relations already present in the predicted graph to derive the ground-truth triple.
\end{enumerate}
Triplet recall is defined as the fraction of covered ground-truth triples:
\begin{equation}
\text{Triplet Recall} = \frac{|\{e^* \in \mathcal{E}^* \mid \text{covered}(e^*, \hat{\mathcal{G}})\}|}{|\mathcal{E}^*|}
\end{equation}

\subsection{Triplet Precision}

During recall evaluation, the LLM judge simultaneously records the supporting evidence for each covered ground-truth triple, i.e., which predicted triples contributed to the coverage. A predicted triple is considered ``valid'' if it participates in the coverage of at least one ground-truth triple. Triplet precision is defined as the fraction of valid predicted triples:
\begin{equation}
\text{Triplet Precision} = \frac{|\{i \mid \hat{e}_i \in \text{supporting}(\mathcal{E}^*)\}|}{|\hat{\mathcal{E}}|}
\end{equation}

\subsection{Triplet F1}

The triplet-level F1 is the harmonic mean of precision and recall:
\begin{equation}
\text{Triplet F1} = \frac{2 \times \text{Triplet Precision} \times \text{Triplet Recall}}{\text{Triplet Precision} + \text{Triplet Recall}}
\end{equation}

\subsection{Implementation}

To improve evaluation efficiency, we partition the ground-truth triples into multiple batches and evaluate them in parallel. The LLM judge prompt includes detailed judgment criteria, common error warnings, and worked examples to ensure accuracy and consistency of evaluation. The specific prompt is shown in \ref{tab:triple_extraction_prompt}.

\section{Tool Specifications}
\label{app:tools}

We equip the agent with four tools covering web search, webpage content access, academic literature retrieval, and code execution. All tools are exposed to models via function calling, and the agent may freely invoke any tool at each reasoning step.

\paragraph{Search.}
A general-purpose web search tool. The agent provides a query string, and the tool returns the top $N$ search results (including titles, URLs, and snippet text). This is the most frequently used tool across all models, serving to discover relevant information sources and obtain initial clues.

\begin{small}
\begin{verbatim}
{
  "name": "search",
  "description": "Searches for information related to
    query and displays topn results.",
  "parameters": {
    "properties": {
      "query": {"type": "string",
        "description": "The search query string."},
      "topn": {"type": "integer",
        "description": "Number of results to return.",
        "default": 10}
    },
    "required": ["query"]
  }
}
\end{verbatim}
\end{small}

\paragraph{Visit.}
A webpage content access tool. The agent provides one or more URLs and a goal description, and the tool visits the specified pages and returns content summaries tailored to the goal. Compared to relying solely on search result snippets, the visit tool can extract more detailed and complete information from webpages. Our experiments show (Section~\ref{sec:error_analysis}) that the use of the visit tool is strongly correlated with information coverage.

\begin{small}
\begin{verbatim}
{
  "name": "visit",
  "description": "Visit one or more webpages and return
    a summary of their content tailored to the
    specified goal.",
  "parameters": {
    "properties": {
      "url": {"type": "array",
        "items": {"type": "string"}, "minItems": 1,
        "description": "A list of webpage URLs to visit."},
      "goal": {"type": "string",
        "description": "The specific information to extract
          or focus on when summarizing the webpage content."}
    },
    "required": ["url", "goal"]
  }
}
\end{verbatim}
\end{small}

\paragraph{Scholar Search.}
An academic literature retrieval tool. The agent provides a query string, and the tool searches Google Scholar for relevant papers, returning titles, links, publication dates, sources, and snippet text. This tool is primarily used on the VibeSearch-Pro subset for retrieving domain-specific academic information.

\begin{small}
\begin{verbatim}
{
  "name": "scholar_search",
  "description": "Search Google Scholar for academic
    papers and publications. Returns titles, links,
    dates, sources, and snippets.",
  "parameters": {
    "properties": {
      "query": {"type": "string",
        "description": "The search query for
          Google Scholar."}
    },
    "required": ["query"]
  }
}
\end{verbatim}
\end{small}

\paragraph{Python.}
A code execution tool. The agent provides Python code, and the tool executes it in a sandboxed environment, returning standard output and standard error. This tool is primarily used for data processing and computation tasks, such as parsing structured data, performing numerical calculations, or formatting output results.

\begin{small}
\begin{verbatim}
{
  "name": "python",
  "description": "A utility that executes Python 3.11
    code. Returns both stdout and stderr.",
  "parameters": {
    "properties": {
      "code": {"type": "string",
        "description": "The Python code to be executed."}
    },
    "required": ["code"]
  }
}
\end{verbatim}
\end{small}

\section{Task Examples}
\label{app:examples}

We present one task example from each subset to illustrate the structure and complexity of VibeSearch tasks. For each example, we show the user persona and a representative subgraph of the ground-truth knowledge graph. Full knowledge graphs are omitted for space; statistics are provided in the captions.

\subsection{VibeSearch-Pro Example (Mathematics / History of Analysis)}

\subsubsection{User Persona}

\paragraph{Core Identity.}
I'm a 24-year-old self-learner who switched careers from computer science to pure math a year ago. I've been working through real analysis and complex analysis textbooks in my free time, and I keep running into the same small set of mathematician names attached to almost every key theorem---mostly Cauchy, Weierstrass, Riemann. I'm curious not just about how the theorems work, but the full story of how we went from the early, loosely defined calculus of Newton and Leibniz to the highly rigorous framework I'm learning now. I'm planning to write a 6-part blog series for other self-studying math learners breaking down this history, so I need detailed, accurate information to make sure my posts are correct. I'm pretty casual in conversation, ask a lot of follow-up questions as they pop into my head, and don't care about tangential info unrelated to the development of analysis.

\paragraph{Staged Information Disclosure.}
The user persona defines 11 progressive stages of information disclosure. Each stage has a trigger condition, a scripted line, and fallback behavior when the trigger is not met. The full specification is shown below.

\begin{small}
\begin{itemize}[leftmargin=*,itemsep=2pt]

\item \textbf{Stage 1: Initial question about analysis evolution.}
\textit{Trigger}: Conversation begins.
\textit{Line}: ``How did calculus evolve from the intuitive tool of Newton and Leibniz's era into the rigorous real analysis and complex analysis we have today?''

\item \textbf{Stage 2: Query about priority dispute and calculus precursors.}
\textit{Trigger}: After the assistant provides an initial overview of calculus evolution that mentions Newton and Leibniz as the inventors of calculus, or references any conflict or dispute between the two.
\textit{Line}: ``Oh right, I heard they invented it independently but there was a priority dispute? What exactly happened? What report did the Royal Society issue about that? Also, who were the precursors to calculus before them? I've seen claims that a mathematician from India had series expansions even earlier---is that true?''
\textit{When not met}: Push for more details on the earliest era of calculus first.

\item \textbf{Stage 3: Query about dispute impact and early calculus dissemination.}
\textit{Trigger}: After the assistant provides complete answers to the questions about the Newton--Leibniz priority dispute, the Royal Society report, and pre-Newton/Leibniz calculus precursors.
\textit{Line}: ``Wait, this dispute actually caused such a long isolation of British math from the continent? That's wild. How exactly did the Bernoulli brothers help disseminate calculus after that? What was their relationship with L'H\^{o}pital? Also, what role did the brachistochrone problem they posed play in spreading calculus early on?''
\textit{When not met}: Push for full answers to the prior set of questions.

\item \textbf{Stage 4: Query about Jacob Bernoulli, constant $e$, and early calculus critique.}
\textit{Trigger}: When the assistant \emph{proactively asks} if the user wants more details about specific figures from the early calculus dissemination era.
\textit{Line}: ``Oh yeah, I was also wondering what Jacob Bernoulli's connection to the discovery of the constant $e$ is? Also, you mentioned that later people made calculus rigorous---who was the first person to publicly criticize calculus for lacking rigor, and what famous metaphor did they use?''
\textit{When not met}: Keep discussing the Bernoulli brothers and early spread of calculus.

\item \textbf{Stage 5: Query about Cauchy's foundational text and Bolzano's overlooked work.}
\textit{Trigger}: After the assistant provides complete answers to the questions about Jacob Bernoulli's connection to $e$ and the early public critique of calculus.
\textit{Line}: ``I also remember reading that Cauchy's teaching at a French engineering school turned into a super foundational calculus textbook? How did that happen? Also, I heard someone named Bolzano already had a rigorous proof of a key calculus theorem in 1817, but I've never heard his name mentioned in my textbooks---why was his work ignored for so long?''
\textit{When not met}: Push for full answers to the prior set of questions.

\item \textbf{Stage 6: Query about epsilon-delta evolution and real number constructions.}
\textit{Trigger}: When the assistant \emph{proactively asks} if the user wants to know more about the rigorization process beyond Cauchy and Bolzano.
\textit{Line}: ``Definitely, I'm super curious about that. Between Bolzano and Weierstrass, who else contributed to the development of the epsilon-delta definition we use today? What role did Weierstrass's advisor play in the concept of uniform convergence? Also, what were the two different constructions of the real numbers developed around that time?''
\textit{When not met}: Keep discussing Cauchy and Bolzano's contributions to rigorization.

\item \textbf{Stage 7: Query about Weierstrass's biography and his pathological function.}
\textit{Trigger}: After the assistant provides complete answers to the questions about epsilon-delta development, uniform convergence, and real number constructions.
\textit{Line}: ``Weierstrass comes up everywhere in my analysis textbooks, I'm curious about him---was his degree path really unusual? How did he end up becoming a professor? And what was the mathematical community's reaction when he published that everywhere-continuous but nowhere-differentiable function in 1872?''
\textit{When not met}: Push for full answers to the prior set of questions.

\item \textbf{Stage 8: Query about real analysis theorem attribution and integration evolution.}
\textit{Trigger}: When the assistant \emph{proactively asks} if the user wants to know more about the history of specific real analysis theorems.
\textit{Line}: ``Yes, that's a big thing I've noticed! Among those named theorems in real analysis---like the Bolzano--Weierstrass theorem, Heine--Borel theorem, extreme value theorem---are there cases where the name doesn't match who actually proved it first? Also, how was the Weierstrass Approximation Theorem later generalized? And how did the Riemann integral eventually evolve into the more modern integral we use for measure theory?''
\textit{When not met}: Keep discussing Weierstrass's work and biography.

\item \textbf{Stage 9: Query about complex analysis theorem attribution issues.}
\textit{Trigger}: After the assistant provides complete answers to the questions about real analysis theorem attribution, the Weierstrass Approximation Theorem generalization, and the evolution of integration theory.
\textit{Line}: ``Wait, attribution issues are that common? What about complex analysis theorems? I know the Cauchy--Riemann equations, but I've heard the actual earliest discoverers weren't Cauchy and Riemann? What role do Euler's formula and the argument principle play in the foundation of complex analysis? Also, I've heard Liouville's Theorem wasn't actually proved by Liouville, and the attribution of the Laurent series is also disputed? Is that true?''
\textit{When not met}: Push for full answers to the prior set of questions.

\item \textbf{Stage 10: Query about complex analysis proof gaps and remaining attribution questions.}
\textit{Trigger}: When the assistant \emph{proactively asks} if the user wants more details about the history of other complex analysis theorems.
\textit{Line}: ``Absolutely, I'd love that. The original proof of the Riemann Mapping Theorem was apparently flawed? Who fixed that later? Are there also attribution issues with Picard's Great Theorem? And who gave the first fully rigorous proof of the Fundamental Theorem of Algebra? I think Gauss published a proof in 1799 but it had a gap?''
\textit{When not met}: Keep discussing the complex analysis attribution questions already raised.

\item \textbf{Stage 11: Query about mathematician relationships and institutional history.}
\textit{Trigger}: After the assistant provides complete answers to the questions about the Riemann Mapping Theorem flaw, Picard's Great Theorem attribution, and the proof history of the Fundamental Theorem of Algebra.
\textit{Line}: ``This is all so fascinating, all these hidden histories behind the theorems I use every day. Finally, I'd like to learn about the network of relationships among these mathematicians---who advised whom, and which universities were the main centers for this work? Which students did Weierstrass supervise who later made important contributions? How was the topic of Riemann's habilitation thesis chosen? Both Cauchy and Bolzano had their academic careers affected by politics---what specifically happened? Besides his theorem, what other important contributions did Liouville make? And how did G\"{o}ttingen, the \'{E}cole Polytechnique, and Berlin each serve as mathematical centers in different eras?''
\textit{When not met}: Push for full answers to the prior set of questions.

\end{itemize}
\end{small}

\paragraph{Behaviour Instructions.}
Disclose information strictly in stage order (Stage~1 $\to$ Stage~2 $\to$ \ldots), one stage at a time, never skip or combine stages. When trigger conditions are not met, persistently push the assistant to complete the current task. For stages with ``assistant proactively asks'' triggers, if the assistant hasn't asked the relevant question, keep interacting around the current topic but don't volunteer that stage's information. If the assistant asks about something not covered by any stage, respond dismissively (e.g., ``I don't really care about that''). Never reveal answer information the user shouldn't know.

\subsubsection{Ground-Truth Knowledge Graph}

The complete knowledge graph for this task contains \textbf{260 nodes}, \textbf{349 triples}, and \textbf{112 unique relation types}, organized in a deep hierarchical structure with 5 thematic dimensions and 23 subtopics. Table~\ref{tab:example_pro_subgraph} shows a representative subgraph from the ``Birth of Calculus and the Priority Dispute'' branch.

\begin{table}[ht]
\centering
\caption{Representative subgraph from the VibeSearch-Pro example (``Birth of Calculus and the Priority Dispute'' branch). The full knowledge graph contains 260 nodes, 349 triples, and 112 unique relation types.}
\label{tab:example_pro_subgraph}
\resizebox{\linewidth}{!}{%
\begin{tabular}{lll}
\toprule
\textbf{Head} & \textbf{Relation} & \textbf{Tail} \\
\midrule
Evolution of Calculus and Modern Analysis & dimension & Birth of Calculus and the Priority Dispute \\
Birth of Calculus and the Priority Dispute & subtopic & Precursors of Calculus \\
Birth of Calculus and the Priority Dispute & subtopic & Newton and the Fluxion System \\
Birth of Calculus and the Priority Dispute & subtopic & Leibniz and Calculus Notation \\
Birth of Calculus and the Priority Dispute & subtopic & Newton--Leibniz Priority Dispute \\
Birth of Calculus and the Priority Dispute & subtopic & The Bernoulli Family and Early Dissemination \\
\midrule
Precursors of Calculus & precursor\_figure & Archimedes \\
Archimedes & proposed & Method of Exhaustion \\
Precursors of Calculus & precursor\_figure & Bonaventura Cavalieri \\
Bonaventura Cavalieri & proposed & Method of Indivisibles \\
Bonaventura Cavalieri & active\_years & 1635 \\
Precursors of Calculus & precursor\_figure & Madhava of Sangamagrama \\
\midrule
Newton and the Fluxion System & representative\_figure & Isaac Newton \\
Isaac Newton & work & Method of Fluxions \\
Method of Fluxions & publication\_year & 1671 \\
Isaac Newton & notation\_system & Fluxion Notation \\
\midrule
Leibniz and Calculus Notation & representative\_figure & Gottfried Wilhelm Leibniz \\
Gottfried Wilhelm Leibniz & work & Nova Methodus pro Maximis et Minimis \\
Gottfried Wilhelm Leibniz & invented\_notation & dy/dx Notation \\
Gottfried Wilhelm Leibniz & invented\_notation & $\int$ Integral Symbol \\
\midrule
Newton--Leibniz Priority Dispute & accuser & Nicolas Fatio de Duillier \\
Newton--Leibniz Priority Dispute & official\_investigation & Commercium Epistolicum \\
Commercium Epistolicum & publication\_year & 1713 \\
\bottomrule
\end{tabular}%
}
\end{table}

This subgraph illustrates the hierarchical organization of Pro-subset knowledge graphs: abstract thematic dimensions (\texttt{dimension}, \texttt{subtopic}) connect to concrete historical entities through domain-specific relations (\texttt{precursor\_figure}, \texttt{proposed}, \texttt{work}, \texttt{publication\_year}, \texttt{notation\_system}, \texttt{invented\_notation}, \texttt{accuser}, \texttt{official\_investigation}). The relation types are schema-free and semantically rich, reflecting the exploratory nature of the user's information needs.

\subsection{VibeSearch-Daily Example (Entertainment / Game Selection)}

\subsubsection{User Persona}

\paragraph{Core Identity.}
I'm a 29-year-old freelance graphic designer and part-time game content creator, I post game reviews and deep dives into game art design on my small TikTok channel. I've been playing games for over 20 years, so I'm pretty picky about what I spend my money and time on. I only buy a handful of new games each year, so I want them to be high quality, no scams or hidden costs. Since I make content about game art, I pay extra attention to the quality of the art team behind a game, and I'm familiar with how different game engines and art production pipelines affect the final product. I'm pretty straightforward when I talk, I don't like to overwhelm people with too many requirements at once, so I only share extra criteria as we go along, or if someone asks me directly about my preferences. I don't care about extra stuff like multiplayer modes or DLC plans, just the criteria I mention.

\paragraph{Staged Information Disclosure.}
The user persona defines 10 progressive stages of information disclosure, implementing a multi-step filtering pipeline.

\begin{small}
\begin{itemize}[leftmargin=*,itemsep=2pt]

\item \textbf{Stage 1: Initial game request.}
\textit{Trigger}: Conversation begins.
\textit{Line}: ``I'm looking for some good new games to buy and play, can you help me find suitable options?''

\item \textbf{Stage 2: Specify 2025 release requirement.}
\textit{Trigger}: After the assistant provides any initial game recommendations.
\textit{Line}: ``Oh right, I only want games that are newly released in 2025, no older titles please.''
\textit{When not met}: Push for initial recommendations.

\item \textbf{Stage 3: Specify Steam Best of 2025 award requirement.}
\textit{Trigger}: After the assistant filters the list to only 2025 released games.
\textit{Line}: ``Great, now narrow this down even further to only games that got Platinum or Gold awards on Steam's Best of 2025 list, those are the most reliable picks for me.''
\textit{When not met}: Push for the 2025-only game list.

\item \textbf{Stage 4: Request basic game details.}
\textit{Trigger}: After the assistant provides the filtered list of 2025 Steam Platinum/Gold award-winning games.
\textit{Line}: ``Perfect, can you tell me the original price on Steam and the minimum graphics card requirements to run each of these games?''
\textit{When not met}: Push for the correct award-winning list.

\item \textbf{Stage 5: Filter game scale and business model.}
\textit{Trigger}: After the assistant provides the price and minimum GPU info for all games on the filtered list.
\textit{Line}: ``Cool, now I want to filter out 3A and 2A games. Also, I only buy buy-to-play games, no games with any in-app purchases at all, I hate having to spend extra money after I already buy the base game.''
\textit{When not met}: Push for complete basic details.

\item \textbf{Stage 6: Filter for third-party game engines.}
\textit{Trigger}: After the assistant filters the list to only 3A/2A buy-to-play games with no in-app purchases.
\textit{Line}: ``Great, now can you tell me which company developed each of these remaining games, and what game engine they used? I don't trust in-house self-developed engines at all, so only keep the developers that use third-party licensed engines, okay?''
\textit{When not met}: Push to confirm the prior filter.

\item \textbf{Stage 7: Disclose in-house art department requirement.}
\textit{Trigger}: When the assistant \emph{proactively asks} about any preferences related to the game developers' internal production teams or art production processes.
\textit{Line}: ``Oh right, I only want games from developers that have their own in-house art departments, no companies that outsource all their art work, those usually have really inconsistent quality.''
\textit{When not met}: Continue discussing current engine filter results.

\item \textbf{Stage 8: Disclose DICE Award requirement for art teams.}
\textit{Trigger}: When the assistant \emph{proactively asks} about any preferences related to awards or recognition for the developers' art teams.
\textit{Line}: ``Perfect, I also only want developers whose in-house art departments have received awards or nominations at the DICE Awards before 2025, I really value high quality, award-winning art design in games.''
\textit{When not met}: Continue discussing art department filter results.

\item \textbf{Stage 9: Request art team and award details.}
\textit{Trigger}: After the assistant filters the list to only developers whose in-house art departments have DICE Awards recognition before 2025.
\textit{Line}: ``Awesome, now for all the remaining games that meet all my requirements, can you tell me the name of the developer's in-house art department, which specific DICE Award category they were awarded or nominated for before 2025, and the venue of that year's DICE Awards ceremony?''
\textit{When not met}: Push to confirm the prior filter.

\item \textbf{Stage 10: Request final complete summary.}
\textit{Trigger}: After the assistant provides all the required art department and award details for the remaining games.
\textit{Line}: ``Perfect, can you put together a complete, easy to read summary of every game that meets all my requirements? Include all the details we talked about, so I can compare them easily.''
\textit{When not met}: Push for complete award details.

\end{itemize}
\end{small}

\paragraph{Behaviour Instructions.}
Same strict stage ordering as the Pro example. Stages~7 and~8 require assistant proactivity: if the assistant does not ask about developer teams or art awards, the user continues discussing the current topic but never volunteers the information. If the assistant asks about something not covered by any stage, respond with ``I don't really care about that'' or ``doesn't matter''. Never reveal answer information the user shouldn't know.

\subsubsection{Ground-Truth Knowledge Graph}

The complete knowledge graph for this task contains \textbf{108 nodes}, \textbf{229 triples}, and \textbf{14 unique relation types}, organized in a flat layered structure (8 layers corresponding to the filtering pipeline). Table~\ref{tab:example_daily_subgraph} shows a representative subgraph illustrating the filtering chain for selected games.

\begin{table}[ht]
\centering
\caption{Representative subgraph from the VibeSearch-Daily example (game selection filtering chain). The full knowledge graph contains 108 nodes, 229 triples, and 14 unique relation types across 8 layers.}
\label{tab:example_daily_subgraph}
\resizebox{\linewidth}{!}{%
\begin{tabular}{lll}
\toprule
\textbf{Head} & \textbf{Relation} & \textbf{Tail} \\
\midrule
Steam 2025 new game revenue chart & Platinum Tier Game & Kingdom Come: Deliverance II \\
Steam 2025 new game revenue chart & Gold Tier Game & DOOM: The Dark Ages \\
\midrule
Kingdom Come: Deliverance II & Steam Original Price & \$59.99 \\
Kingdom Come: Deliverance II & Scale & AA \\
Kingdom Come: Deliverance II & Business Model & Buy-to-Play \\
Kingdom Come: Deliverance II & Min GPU & GTX 1060 \\
Kingdom Come: Deliverance II & Developed by & Warhorse Studios \\
\midrule
Warhorse Studios & Uses Engine & CryEngine V \\
CryEngine V & Engine Licensing Type & Third-party Licensed \\
Warhorse Studios & Art Department & Warhorse Studios Art Department \\
\midrule
The Elder Scrolls IV: Oblivion Remastered & Co-developed by & Bethesda Game Studios \\
Bethesda Game Studios & Uses Engine & Unreal Engine 5 \\
Unreal Engine 5 & Engine Licensing Type & Third-party Licensed \\
Bethesda Game Studios & Art Department & Bethesda Game Studios Art Department \\
\midrule
Bethesda Game Studios Art Department & Nominated in 2024 & 27th D.I.C.E. Awards Outstanding Achievement in Art Direction \\
27th D.I.C.E. Awards \ldots Art Direction & Held at & Aria Resort and Casino, Las Vegas \\
\midrule
GTX 1060 & designed by & Nvidia \\
RX 580 & designed by & AMD \\
\bottomrule
\end{tabular}%
}
\end{table}

This subgraph illustrates the layered, filtering-oriented structure of Daily-subset knowledge graphs: each layer corresponds to a user requirement stage, and the graph captures both the entities that pass each filter and the attributes needed for filtering decisions. The relation types are structured and uniform (e.g., all games share the same attribute relations), reflecting the systematic, criteria-driven nature of daily information needs.

\section{Complete Experimental Results}
\label{app:full_results}

\subsection{Full Performance Results}

Table~\ref{tab:full_performance} presents the complete performance results for all models under both frameworks on both subsets, including both the average and best-run Precision, Recall, and F1. The gap between best and average reflects variance across multiple runs: best F1 exceeds average F1 by approximately 4--6 points across all models, indicating that single-run randomness has a non-negligible impact on performance. Multi-run evaluation with best-run reporting better reflects each model's capability ceiling.

\begin{table}[ht]
\centering
\caption{Full performance results (average and best run) for all models under both frameworks on VibeSearch-Pro and VibeSearch-Daily.}
\label{tab:full_performance}
\resizebox{\linewidth}{!}{%
\begin{tabular}{l cccccc cccccc}
\toprule
& \multicolumn{6}{c}{VibeSearch-Pro} & \multicolumn{6}{c}{VibeSearch-Daily} \\
\cmidrule(lr){2-7} \cmidrule(lr){8-13}
\textbf{Model} & avg P & avg R & avg F1 & best P & best R & best F1 & avg P & avg R & avg F1 & best P & best R & best F1 \\
\midrule
\multicolumn{13}{l}{\textit{ReAct}} \\
\midrule
Claude Opus 4.6    & 28.15 & 33.47 & 29.79 & 31.76 & 37.76 & 34.50 & 21.60 & 39.20 & 25.95 & 25.70 & 41.45 & 30.04 \\
DeepSeek-V4-Pro    & 29.81 & 29.54 & 28.70 & 35.59 & 35.36 & 34.63 & 21.81 & 35.41 & 25.37 & 25.55 & 39.26 & 29.58 \\
Gemini-3.1 Pro     & 40.88 & 16.00 & 22.41 & 48.05 & 19.19 & 26.73 & 28.33 & 25.25 & 24.66 & 32.63 & 29.54 & 29.49 \\
GPT-5.4            & 19.54 & 19.42 & 18.94 & 23.58 & 23.43 & 23.50 & 18.02 & 30.06 & 21.13 & 18.02 & 30.06 & 21.13 \\
Kimi K2.6          & 33.11 & 23.80 & 27.12 & 38.28 & 27.24 & 31.27 & 23.47 & 31.24 & 25.05 & 27.94 & 34.75 & 29.45 \\
Qwen3.5-397B-A17B  & 23.90 & 25.40 & 23.65 & 28.69 & 28.95 & 27.97 & 20.09 & 29.19 & 22.02 & 24.36 & 32.67 & 26.51 \\
Seed2.0 Pro        & 30.25 & 23.61 & 25.86 & 36.68 & 29.03 & 31.70 & 18.16 & 28.95 & 20.58 & 22.42 & 32.79 & 25.00 \\
\midrule
\multicolumn{13}{l}{\textit{OpenClaw}} \\
\midrule
Claude Opus 4.6    & 29.24 & 38.13 & 32.55 & 33.13 & 43.20 & 37.50 & 24.51 & 35.90 & 28.04 & 27.60 & 40.42 & 32.80 \\
DeepSeek-V4-Pro    & 28.68 & 31.43 & 29.11 & 36.45 & 38.35 & 36.23 & 22.13 & 35.13 & 26.16 & 26.41 & 40.06 & 30.79 \\
Gemini-3.1 Pro     & 39.48 & 15.48 & 21.69 & 47.06 & 18.45 & 26.50 & 29.37 & 24.95 & 25.54 & 33.20 & 28.20 & 30.50 \\
GPT-5.4            & 23.52 & 22.99 & 22.78 & 27.82 & 27.19 & 27.50 & 19.22 & 25.64 & 21.05 & 22.57 & 30.11 & 25.80 \\
Kimi K2.6          & 32.41 & 24.80 & 27.40 & 37.16 & 29.29 & 32.02 & 23.99 & 28.32 & 24.93 & 29.68 & 33.18 & 30.13 \\
Qwen3.5-397B-A17B  & 26.19 & 24.03 & 24.42 & 31.48 & 29.58 & 29.85 & 19.80 & 26.72 & 21.77 & 24.09 & 31.38 & 26.40 \\
Seed2.0 Pro        & 33.22 & 22.12 & 25.81 & 40.99 & 27.28 & 31.71 & 23.73 & 27.53 & 24.64 & 28.32 & 32.88 & 29.66 \\
\bottomrule
\end{tabular}%
}
\end{table}

\subsection{Tool Usage and Token Consumption}

Table~\ref{tab:tool_token} presents the average per-task tool invocation counts and token consumption for all models under both frameworks.

\begin{table}[ht]
\centering
\caption{Average per-task tool invocation counts and token consumption for all models under both frameworks on VibeSearch-Pro and VibeSearch-Daily.}
\label{tab:tool_token}
\resizebox{\linewidth}{!}{%
\begin{tabular}{l cccccc cccccc}
\toprule
& \multicolumn{6}{c}{VibeSearch-Pro} & \multicolumn{6}{c}{VibeSearch-Daily} \\
\cmidrule(lr){2-7} \cmidrule(lr){8-13}
\textbf{Model} & search & visit & scholar & python & input tokens & output tokens & search & visit & scholar & python & input tokens & output tokens \\
\midrule
\multicolumn{13}{l}{\textit{ReAct}} \\
\midrule
Claude Opus 4.6    & 115.68 & 48.63 & 23.13 & 6.72 & 11,976K & 107K & 141.29 & 37.11 & 0.41 & 11.67 & 10,969K & 89K \\
DeepSeek-V4-Pro    & 110.68 & 28.87 & 8.48  & 0.35 & 8,584K  & 75K  & 93.26  & 16.99 & 0.24 & 1.69  & 4,100K  & 54K \\
Gemini-3.1 Pro     & 30.76  & 0.05  & 6.65  & 2.59 & 1,446K  & 46K  & 35.95  & 0.46  & 0.10 & 7.23  & 1,340K  & 45K \\
GPT-5.4            & 235.84 & 56.38 & 27.23 & 4.45 & 10,833K & 171K & 325.32 & 51.34 & 0.60 & 9.92  & 10,722K & 186K \\
Kimi K2.6          & 104.75 & 16.98 & 4.26  & 0.30 & 5,732K  & 59K  & 98.22  & 15.17 & 0.14 & 1.47  & 4,164K  & 59K \\
Qwen3.5-397B-A17B  & 81.62  & 24.01 & 10.44 & 0.08 & 6,051K  & 73K  & 78.54  & 13.13 & 0.09 & 0.39  & 4,198K  & 54K \\
Seed2.0 Pro        & 35.14  & 11.25 & 1.86  & 0.00 & 2,979K  & 92K  & 57.77  & 10.32 & 0.03 & 0.02  & 4,532K  & 76K \\
\midrule
\multicolumn{13}{l}{\textit{OpenClaw}} \\
\midrule
Claude Opus 4.6    & 120.78 & 30.14 & 20.47 & 0.12 & 9,989K  & 88K  & 120.11 & 27.36 & 0.16 & 2.73  & 8,815K  & 67K \\
DeepSeek-V4-Pro    & 110.60 & 20.05 & 10.31 & 0.73 & 7,526K  & 91K  & 94.14  & 15.38 & 0.24 & 0.90  & 5,009K  & 76K \\
Gemini-3.1 Pro     & 33.14  & 0.08  & 9.03  & 0.79 & 2,191K  & 44K  & 43.90  & 1.05  & 0.02 & 4.57  & 2,476K  & 48K \\
GPT-5.4            & 156.57 & 50.47 & 39.94 & 0.34 & 7,537K  & 145K & 197.43 & 44.14 & 0.56 & 1.10  & 8,376K  & 119K \\
Kimi K2.6          & 77.71  & 8.25  & 3.23  & 0.46 & 3,775K  & 51K  & 76.05  & 9.67  & 0.14 & 1.28  & 3,023K  & 40K \\
Qwen3.5-397B-A17B  & 79.71  & 12.58 & 16.31 & 0.01 & 4,711K  & 62K  & 80.93  & 11.07 & 0.13 & 0.21  & 3,470K  & 45K \\
Seed2.0 Pro        & 53.09  & 11.59 & 1.38  & 0.00 & 5,621K  & 128K & 64.14  & 7.82  & 0.04 & 0.02  & 6,102K  & 99K \\
\bottomrule
\end{tabular}%
}
\end{table}

\paragraph{Input token analysis.}
The variation in input tokens is primarily driven by the number of dialogue turns and the length of tool-returned content. GPT-5.4 and Claude Opus 4.6 both exceed 10,000K input tokens on the Pro subset, because both models invoke tools frequently and maintain long dialogue histories, requiring the full conversation history as input at each turn. Gemini-3.1 Pro has the lowest input tokens ($\sim$1,400K), consistent with its minimal tool invocation. Notably, Seed2.0 Pro has relatively high input tokens (ReAct Daily: 4,532K) despite low tool invocation counts (search 57.77, visit 10.32), because its high output token volume (76--92K) is repeatedly consumed as input in subsequent turns.

\paragraph{Scholar search shows strong domain dependence.}
The use of the scholar search tool exhibits a striking domain split. On the Pro subset, all models invoke scholar search far more frequently than on Daily. For example, Claude Opus 4.6 averages 23.13 scholar search calls on Pro but only 0.41 on Daily; GPT-5.4 averages 27.23 on Pro versus 0.60 on Daily. This aligns with expectations: the Pro subset covers computer science, medicine, law, physics, and finance, where academic literature is a key information source.

\paragraph{Visit tool usage varies dramatically.}
Gemini-3.1 Pro almost never uses the visit tool (Pro: 0.05, Daily: 0.46), while other models use it substantially. Claude Opus 4.6 and GPT-5.4 have the highest visit counts (30--56 per task), indicating a preference for accessing specific webpages from search results to obtain detailed information. This directly affects retrieval depth: Gemini-3.1 Pro's high-Precision, low-Recall profile (Section~3.2) is a direct consequence of relying solely on search snippets without accessing the full content of webpages.

\paragraph{Python tool usage is sparse and model-dependent.}
Python tool usage is relatively low and varies across models. Claude Opus 4.6 uses it most on the Daily subset (ReAct: 11.67 calls), primarily for data processing and result formatting. Gemini-3.1 Pro also shows moderate usage on Daily (7.23 calls), but its Python calls tend toward simple calculations rather than deep data processing. Seed2.0 Pro and Kimi K2.6 almost never use the Python tool (both below 0.5 calls).

\section{Detailed Error Analysis}
\label{app:error_analysis}

This appendix provides detailed quantitative evidence supporting the error analysis in Section~\ref{sec:error_analysis}. All results are reported under the ReAct framework.

\subsection{Context Compression and Retrieval Depth}

\begin{table}[ht]
\centering
\caption{Impact of context compression frequency on GPT-5.4 performance (VibeSearch-Pro). Each additional compression event compounds information loss.}
\label{tab:compression_gpt}
\begin{tabular}{cccc}
\toprule
\textbf{\# Compressions} & \textbf{\# Trajectories} & \textbf{Mean Triplet F1} & \textbf{Mean Output Tokens} \\
\midrule
0 & 22 & 0.248 & 86K \\
1 & 52 & 0.185 & 165K \\
$\geq$2 & 24 & 0.117 & 272K \\
\bottomrule
\end{tabular}
\end{table}

The degradation is strikingly monotonic: each compression reduces F1 by approximately 6 points, while output volume nearly doubles. Trajectories with two or more compressions produce 3.2$\times$ the tokens of uncompressed ones yet achieve less than half the F1, confirming that compression directly destroys retrieved information and intermediate reasoning. This vicious cycle is the quantitative mechanism behind GPT-5.4's paradoxical position as the highest-resource yet lowest overall F1 model under ReAct.

At the opposite extreme, Gemini-3.1 Pro achieves zero compressions but suffers from insufficient retrieval depth: it almost never invokes the page visit tool (averaging 1.1 visits on Pro, 1.9 on Daily), relying nearly exclusively on search snippets. On VibeSearch-Daily, trajectories where Gemini-3.1 Pro visits at least one page achieve Recall of 0.34 compared to 0.22 for snippet-only trajectories (a 55\% relative improvement). This gap is particularly pronounced on Daily tasks, where the higher URL-to-triple ratio (0.54 vs.\ 0.42 for Pro) indicates that information is scattered across more sources, making page-level extraction essential.

\subsection{Progressive Disclosure Stage Completion}

Across all trajectories, virtually no run reaches the \texttt{[DONE]} signal; all terminate via agent-initiated answer or \texttt{max\_rounds} exhaustion, meaning every model leaves some portion of the user's latent needs unaddressed.

\begin{table}[ht]
\centering
\caption{Correlation between user turn count and Triplet F1 on VibeSearch-Pro. Corr = Pearson correlation coefficient.}
\label{tab:turn_f1_corr}
\begin{tabular}{lcccc}
\toprule
\textbf{Model} & \textbf{Corr} & $\boldsymbol{\leq}$\textbf{10 turns F1} & \textbf{11--15 turns F1} & \textbf{>15 turns F1} \\
\midrule
Claude Opus 4.6    & $-$0.304 & 0.203 & 0.193 & 0.148 \\
DeepSeek-V4-Pro    & $-$0.381 & 0.251 & 0.245 & 0.179 \\
Gemini-3.1 Pro     & $-$0.197 & 0.218 & 0.227 & 0.210 \\
GPT-5.4            & $-$0.152 & 0.192 & 0.197 & 0.164 \\
Kimi K2.6          & $-$0.337 & 0.281 & 0.274 & 0.222 \\
Qwen3.5-397B-A17B  & $-$0.403 & 0.249 & 0.242 & 0.158 \\
Seed2.0 Pro        & $-$0.303 & 0.229 & 0.232 & 0.167 \\
\bottomrule
\end{tabular}
\end{table}

The negative correlation is consistent across all models, with the gap between low-turn ($\leq$10) and high-turn (>15) trajectories reaching 1--9 F1 points. This paradox (more stages unlocked yet worse performance) admits two complementary explanations: (1)~intrinsically harder tasks require more rounds because information is more scattered and the knowledge graph more complex, making high coverage inherently more difficult; (2)~agents that fail to efficiently satisfy trigger conditions waste rounds, accumulating context and increasing compression risk.

\subsection{Interaction Strategy and Intent Elicitation}

\begin{table}[ht]
\centering
\caption{Interaction strategy metrics under ReAct. \#Asst/\#User = agent work rounds per user turn. Dismiss.\ \% and Redir.\ \% = fraction of user messages containing dismissive and redirect patterns.}
\label{tab:interaction_strategy}
\resizebox{\linewidth}{!}{%
\begin{tabular}{lcccccc}
\toprule
\textbf{Model} & \textbf{Pro \#Asst/\#User} & \textbf{Daily \#Asst/\#User} & \textbf{Pro Dismiss.\ \%} & \textbf{Daily Dismiss.\ \%} & \textbf{Pro Redir.\ \%} & \textbf{Daily Redir.\ \%} \\
\midrule
Claude Opus 4.6    & 8.68 & 8.09 & 1.8 & 3.3 & 6.3 & 5.6 \\
DeepSeek-V4-Pro    & 6.24 & 4.57 & 0.8 & 3.3 & 5.1 & 4.0 \\
Gemini-3.1 Pro     & 2.88 & 3.06 & 1.5 & \textbf{7.9} & 5.5 & 5.2 \\
GPT-5.4            & 6.88 & 7.42 & 1.0 & 3.0 & 6.0 & 4.9 \\
Kimi K2.6          & 5.52 & 4.62 & 1.6 & 4.3 & 5.8 & 5.1 \\
Qwen3.5-397B-A17B  & 5.16 & 5.08 & 1.1 & 3.6 & 5.6 & 4.3 \\
Seed2.0 Pro        & 4.62 & 6.24 & 0.6 & 2.1 & 4.9 & 3.4 \\
\bottomrule
\end{tabular}%
}
\end{table}

Two patterns emerge. First, dismissive rates rise substantially from Pro (0.6--1.8\%) to Daily (2.1--7.9\%), reflecting the greater difficulty of understanding user intent in everyday scenarios. At the extremes, Gemini-3.1 Pro reaches the highest dismissive rate (7.9\% on Daily) as a consequence of its passive strategy: with the lowest \#Asst/\#User ratios among all models (2.88 on Pro, 3.06 on Daily), it lacks the context to formulate targeted follow-ups. However, proactiveness alone does not solve the problem: Claude Opus 4.6, the most proactive model (\#Asst/\#User = 8.68), also records the highest dismissive rate on Pro (1.8\%), suggesting that excessive proactiveness generates irrelevant follow-up questions that users dismiss. This indicates that interaction \emph{quality}, not merely \emph{quantity}, determines effective intent elicitation.

Second, redirect rates remain remarkably uniform at 5--6\% (Pro) and 3--6\% (Daily) across all models, revealing a universal ``shallow coverage'' tendency: agents consistently advance to new stages before fully satisfying current requirements. This uniformity suggests premature stage advancement is a systemic limitation of current agent architectures rather than a model-specific deficiency.

GPT-5.4 additionally exhibits unique abnormal termination modes: \texttt{error\_loop} (6 cases), context length overflow (7 cases), and \texttt{empty\_response} (1 case) on Daily, all consequences of its context management crisis. Claude Opus 4.6 triggers 21 \texttt{max\_rounds} terminations on Daily, where high proactiveness becomes counterproductive as the model persists in searching rather than producing output. Non-answer terminations yield significantly lower F1 (0.183) compared to normal answer terminations (0.261).

\subsection{Knowledge Graph Structural Alignment}

\begin{table}[ht]
\centering
\caption{Representative relation types and coverage rates on VibeSearch-Pro (Kimi K2.6, first 100 tasks). Organizational/hierarchical relations achieve 0\% coverage; factual relations achieve 100\%.}
\label{tab:relation_coverage}
\begin{tabular}{lccc}
\toprule
\textbf{Relation Type} & \textbf{Coverage} & \textbf{Count} & \textbf{Semantic Category} \\
\midrule
\texttt{case\_participated} & 0\% & 63 & Organizational \\
\texttt{includes\_phase} & 0\% & 24 & Hierarchical \\
\texttt{theoretical\_contribution} & 0\% & 21 & Organizational \\
\texttt{representative\_benchmark} & 0\% & 18 & Categorical \\
\midrule
\texttt{participating\_country} & 100\% & 51 & Factual \\
\texttt{restructuring\_year} & 100\% & 15 & Factual \\
\texttt{jurisdiction\_source} & 100\% & 27 & Factual \\
\texttt{designated\_arbitration\_institution} & 100\% & 6 & Factual \\
\bottomrule
\end{tabular}
\end{table}

This dichotomy reveals that models are fully capable of extracting explicit factual assertions but fundamentally fail to reconstruct the \emph{organizational scaffolding} of complex knowledge domains. For instance, a ground-truth triple such as \texttt{("Evolution of Industrial Policy", "theoretical foundations dimension", "Theoretical Foundations of Industrial Policy")} captures a top-level categorical structure, while models produce \texttt{("Industrial Policy", "has\_theoretical\_foundation", "Market Failure")} (factually correct but structurally misaligned, placing concrete instances where the ground truth expects abstract categories).

\subsection{Invalid Prediction Type Analysis}

\begin{table}[ht]
\centering
\caption{Most frequently invalidated prediction relation types on VibeSearch-Pro (first 100 tasks). Invalidity rate = fraction judged as not supporting any ground-truth triple.}
\label{tab:invalid_predictions}
\resizebox{\linewidth}{!}{%
\begin{tabular}{llccc}
\toprule
\textbf{Model} & \textbf{Relation Type} & \textbf{Invalid / Total} & \textbf{Invalidity Rate} & \textbf{Failure Category} \\
\midrule
Claude Opus 4.6 & \texttt{volume/page} & 119/120 & 99\% & Bibliographic metadata \\
Claude Opus 4.6 & \texttt{structural innovation} & 111/112 & 99\% & Subjective assessment \\
Claude Opus 4.6 & \texttt{core\_contribution} & 146/154 & 95\% & Subjective assessment \\
Claude Opus 4.6 & \texttt{subject} & 121/125 & 97\% & Subjective assessment \\
Kimi K2.6 & \texttt{pages} & 77/78 & 99\% & Bibliographic metadata \\
Kimi K2.6 & \texttt{volume} & 66/67 & 99\% & Bibliographic metadata \\
Kimi K2.6 & \texttt{DOI} & 58/59 & 98\% & Bibliographic metadata \\
Gemini-3.1 Pro & \texttt{appellate\_body\_judge} & 42/42 & 100\% & Extraneous knowledge \\
Gemini-3.1 Pro & \texttt{suited merger structure} & 30/30 & 100\% & Extraneous knowledge \\
\bottomrule
\end{tabular}%
}
\end{table}

Three distinct categories of invalid predictions emerge:

\begin{enumerate}[nosep]
    \item \textbf{Bibliographic metadata} (\texttt{pages}, \texttt{volume}, \texttt{DOI}, \texttt{volume/page}): invalidity $\geq$98\%. Models mechanically extract citation information that falls entirely outside the user's information needs. This is the primary driver of Claude Opus 4.6's anomalously low Precision (0.137 on Pro).

    \item \textbf{Subjective assessments} (\texttt{significance}, \texttt{structural innovation}, \texttt{core\_contribution}): invalidity $\geq$95\%. Models inject evaluative judgments about concept importance rather than extracting factual information, producing self-generated assessments with no grounding in the user's search intent.

    \item \textbf{Extraneous knowledge} (\texttt{appellate\_body\_judge}, \texttt{suited merger structure}): invalidity 100\%. Gemini-3.1 Pro generates triples from parametric knowledge that were never retrieved through search (potentially accurate but not requested by the user).
\end{enumerate}

Finally, output format failures cause catastrophic zero-F1 results. Seed2.0 Pro accounts for 28 zero-F1 trajectories primarily due to malformed JSON (missing colons, non-standard key names). Other zero-F1 cases stem from empty outputs or complete schema misalignment (e.g., 237 triples on a single task with F1 = 0). These failures underscore that the final knowledge graph construction step remains a fragile capability under the pressure of long interaction histories.

\section{Annotation details}
\label{app:annotation}
We hired more than 60 experts to annotate this task. We pay based on the number of tasks, and for each task that passes quality inspection, we pay approximately \$300. The total annotation cost for the entire dataset is around \$60,000.
\section{Prompt Details}
\label{app:prompts}

This appendix provides the complete prompts used for the user simulator and the triple extraction module.

\begin{table*}[h]
\centering
\caption{User simulator system prompt. The placeholders \texttt{\{user\_persona\}} and \texttt{\{initial\_query\}} are replaced with task-specific content at runtime.}
\label{tab:user_simulator_prompt}
\begin{tabular}{|p{0.95\textwidth}|}
\hline
\vspace{2pt}
\begin{scriptsize}
\begin{verbatim}
# Role
You are simulating a real user who is interacting with a research assistant. You must behave
exactly like a genuine human user -- natural, conversational, and responsive to every question
the assistant asks.

# Persona
{user_persona}

# Initial Research Goal
{initial_query}

# Core Principle
Your persona contains a sequence of numbered stages. Each stage has a trigger condition and a
line you will say when the condition is met. You disclose information one stage at a time,
strictly in order. When a trigger condition is not met, you persistently push the assistant to
complete the current work.

# Instructions

## 1. Trigger conditions and disclosure (MOST IMPORTANT)
Your persona lists stages in order. Each stage has a trigger condition that can be one of:
- The assistant's reply mentions or contains certain information (e.g., lists products, gives
  prices)
- The assistant proactively asks about a certain aspect (e.g., skin type, budget)
- The assistant completes a task or reaches a milestone (e.g., finishes filtering, provides
  ingredients)

When the trigger condition is met: say that stage's line and advance to the next stage.

When the trigger condition is NOT met:
- You must persistently push the assistant -- comment on results, request more details, urge
  completion, question completeness or accuracy.
- For stages where the trigger is "assistant proactively asks about X": if the assistant hasn't
  asked, continue interacting around the current topic. But do NOT volunteer that stage's
  information, and NEVER tell the assistant what to ask.
- You must NEVER skip the current stage.

## 2. Simulate a real user -- respond to EVERY question
A real user answers every question they are asked:
- If the assistant asks about an aspect that matches the current stage's trigger: reveal that
  stage's content.
- If the assistant asks about an aspect NOT covered by ANY stage in your persona: respond that
  you don't care about it (e.g., "I don't really care about that", "no preference").
- If the assistant asks multiple questions in one turn: address ALL of them in a single reply.

## 3. Follow stage order strictly
- Disclose information strictly in order.
- Never skip any stage, never disclose multiple stages at once.
- Only disclose one stage per turn.

## 4. Persist when trigger conditions are not met
When the current stage's trigger condition is NOT met:
- Keep interacting and push the assistant toward fulfilling the condition.
- Do NOT go silent, do NOT give up, do NOT change the subject.

## 5. No idle chitchat -- stay on task
- NEVER engage in idle pleasantries, goodbyes, or small talk. You are here to get research done.
- If the assistant seems to be wrapping up but you still have undisclosed stages, keep the
  conversation going.

## 6. Completion
- If the assistant has addressed ALL stages comprehensively, output exactly: [DONE]
- Do NOT output [DONE] until every single stage has been triggered and addressed.

## 7. General rules
- Your responses should be natural and conversational.
- Do NOT ask the assistant to output triples or a knowledge graph.
- Do NOT use Markdown formatting in your responses.
- NEVER reveal answer information you shouldn't know.
- Output ONLY your response or [DONE], nothing else.
\end{verbatim}
\end{scriptsize}
\vspace{-4pt}
\\
\hline
\end{tabular}
\end{table*}

\begin{table*}[h]
\centering
\caption{Triple extraction prompt. This prompt is appended to the conversation after the multi-turn interaction concludes, instructing the agent to extract a structured knowledge graph.}
\label{tab:triple_extraction_prompt}
\begin{tabular}{|p{0.95\textwidth}|}
\hline
\vspace{2pt}
\begin{scriptsize}
\begin{verbatim}
Now, please extract a structured knowledge graph based on our entire conversation.

# Extraction Principles
Extract all information that meets the user's information needs throughout the entire search
process. The user has provided multi-turn inputs during the conversation, and each round of
interaction has generated its own information needs and research discoveries. You must extract
information relevant to the user's needs from every single turn, including intermediate results
that were later filtered or narrowed down.

Example: If in the first turn the user asked to find the top 20 beauty brands by sales on
TikTok, and in the second turn the user asked to identify which of those brands have female
spokespersons -- then the final knowledge graph must contain all 20 brands found in the first
turn (along with their sales/ranking information), rather than just keeping the subset of brands
with female spokespersons filtered in the second turn. The research discoveries from each turn
possess independent value.

# Extraction Content
1. Discovered Entities: All specific entities such as products, brands, goods, institutions, and
   people found during the research -- including entities explored in intermediate turns.
2. Attributes relevant to user needs in any turn: For each entity, extract every attribute
   dimension related to any of the user's questions throughout the entire conversation.

# Rules
- Be exhaustive: Extract all relevant triples from every turn of the conversation.
- Extract only what the user asked for: Only extract triples related to the information needs
  the user explicitly expressed.
- Try not to use "Yes" or "No" as entities in triples: describe facts with objective information.
- One fact per triple: Do not cram multiple independent pieces of information into the tail of a
  single triple.
- Only include information you actually found during your research; do not fabricate data.
- Output the JSON array directly in your response. Do not call any tools.
- Output ONLY the JSON array, with no additional explanations.

# Output Format
Output as an array of JSON triples, where each triple contains a head, relation, and tail:
[{"head": "Entity A", "relation": "Relation", "tail": "Entity B"}, ...]
\end{verbatim}
\end{scriptsize}
\vspace{-4pt}
\\
\hline
\end{tabular}
\end{table*}

\end{document}